\documentclass[10pt,twocolumn,letterpaper]{article}

\usepackage[pagenumbers]{cvpr} 

\let\svthefootnote\thefootnote

\newcommand\blankfootnote[1]{%
  \let\thefootnote\relax\footnotetext{#1}%
  \let\thefootnote\svthefootnote%
}
\let\svfootnote\footnote
\renewcommand\footnote[2][?]{%
  \if\relax#1\relax%
    \blankfootnote{#2}%
  \else%
    \if?#1\svfootnote{#2}\else\svfootnote[#1]{#2}\fi%
  \fi
}

\definecolor{DarkGray}{RGB}{85, 85, 85}

\definecolor{cvprblue}{rgb}{0.21,0.49,0.74}

\usepackage[pagebackref,breaklinks,colorlinks,allcolors=cvprblue]{hyperref}
\usepackage{tikz}
\usepackage{float}
\usepackage{multirow} 
\usepackage{makecell}
\usepackage{colortbl}
\usepackage{longtable}
\usepackage{adjustbox}
\usepackage{multirow}
\usepackage[ruled,linesnumbered]{algorithm2e}

\newcommand{\yellowcircle}[1]{%
    \tikz[baseline=(char.base)]{
        \node[shape=circle,fill=yellow,draw=black,inner sep=1pt] (char) {\scriptsize\textbf{#1}};%
    }%
}

\SetCommentSty{mycommfont}
\SetKwComment{Comment}{// }{ }
\SetKwInput{KwInput}{Input}

\def\paperID{13649} 
\def\confName{CVPR}
\def\confYear{2025}

\title{Rendering-Refined Stable Diffusion for Privacy Compliant Synthetic Data}

\author{
    Kartik Patwari\textsuperscript{*,$\dag$,a}, \
    David Schneider\textsuperscript{*,$\dag$,b}, \
    Xiaoxiao Sun\textsuperscript{c}, \ 
    Chen-Nee Chuah\textsuperscript{a}, \ \\
    Lingjuan Lyu\textsuperscript{d}, \
    Vivek Sharma\textsuperscript{*,$\star$,d} \\[5pt]
    $^a$University of California, Davis \qquad
    $^b$Karlsruhe Institute of Technology \qquad \\[2pt]
    $^c$Australian National University \qquad
    $^d$Sony AI, Sony Research 
}

\begin{document}

\twocolumn[{%
\vspace{-50px}
\renewcommand\twocolumn[1][]{#1}%
\maketitle
\centering
\includegraphics[width=1.0\linewidth]{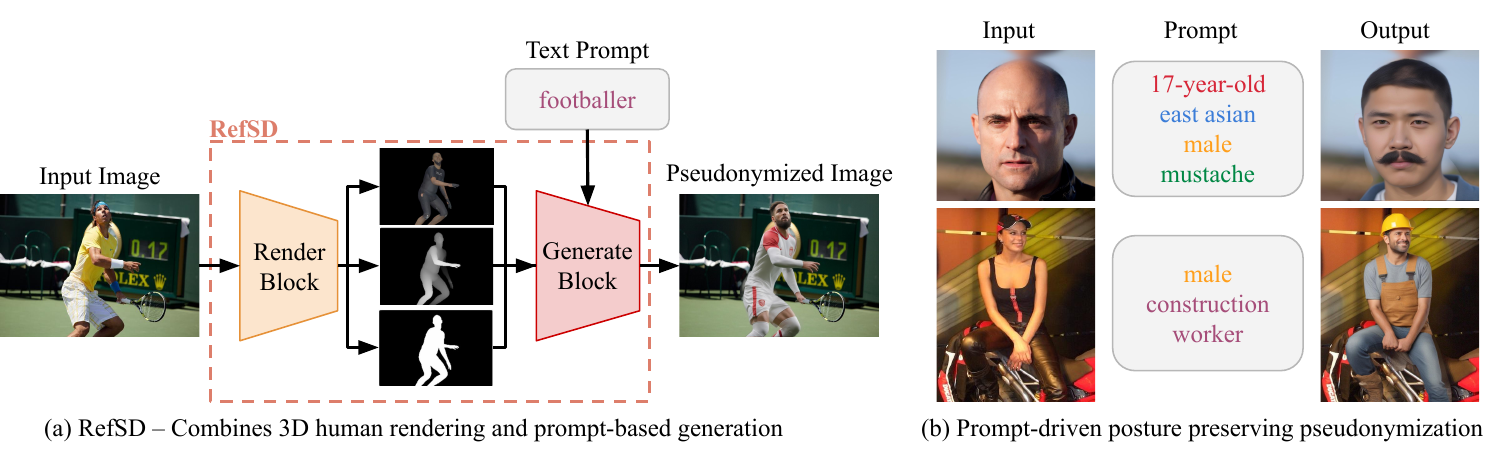}
\vspace{-10px}
\captionof{figure}{\label{fig:intro_fig}
\textbf{Rendering-Refined Stable Diffusion~(RefSD)} pseudonymizes while preserving posture by combining 3D-rendered poses with attribute-driven human generation, as shown in (a). Two examples of pseudonymized images processed by RefSD are shown in (b).}
\vspace{10px}
}]

\footnote[]{\textsuperscript{*}Equal contribution \textsuperscript{$\dag$}Partial Work done while interning at Sony AI \textsuperscript{$\star$}VS started and led the project. Correspondence to: Vivek Sharma $<$viveksharma@sony.com$>$.}

\begin{abstract}
Growing privacy concerns and regulations like GDPR and CCPA necessitate pseudonymization techniques that protect identity in image datasets. However, retaining utility is also essential. Traditional methods like masking and blurring degrade quality and obscure critical context, especially in human-centric images. We introduce \textbf{Rendering-Refined Stable Diffusion (RefSD)}, a pipeline that combines 3D-rendering with Stable Diffusion, enabling prompt-based control over human attributes while preserving posture. Unlike standard diffusion models that fail to retain posture or GANs that lack realism and flexible attribute control, RefSD balances posture preservation, realism, and customization. We also propose HumanGenAI, a framework for human perception and utility evaluation. Human perception assessments reveal attribute-specific strengths and weaknesses of RefSD. Our utility experiments show that models trained on RefSD pseudonymized data outperform those trained on real data in detection tasks, with further performance gains when combining RefSD with real data. For classification tasks, we consistently observe performance improvements when using RefSD data with real data, confirming the utility of our pseudonymized data. 

\end{abstract}
\vspace{-10px}    
\section{Introduction}
\label{sec:intro}

Advances in computer vision have intensified corporate concerns over privacy in image datasets containing personally identifiable information (PII)~\citep{paullada2021data}.
These concerns are particularly pressing in three scenarios: (1) \emph{internal, confidential, or proprietary datasets}, which require strict compliance with data protection regulations even when used internally~\citep{gdpr}; (2) \emph{publicly available data without explicit consent from individuals depicted}, which cannot be legally used without appropriate measures~\citep{ICO2019}; and (3) \emph{public datasets licensed under terms like CC BY 4.0}, where data must be pseudonymized to comply with privacy regulations like General Data Protection Regulation (GDPR)~\citep{gdpr, EDPS2021Anonymisation} and California Consumer Privacy Act (CCPA)~\cite{ccpa}.

\emph{Pseudonymization}, as defined by the GDPR$^1$, involves processing personal data so it cannot be attributed to an individual without additional information~\citep{gdpr}.
In the context of public datasets like OpenImages~\citep{openimages} and Objects365~\citep{objects365}, pseudonymization is essential for commercial use while complying with privacy regulations.

\footnote[]{$^{1}$ GDPR Article 4: \url{https://gdpr-info.eu/art-4-gdpr/}}

However, traditional pseudonymization methods like masking and blurring are obstructive and degrade image utility by obscuring critical context—especially in human-centric applications where interactions are pivotal~\citep{du2019efficient, yang2022study}. Therefore, there is a pressing need for GDPR-compliant pseudonymization techniques that preserve both privacy and data utility. Our work addresses this challenge by developing methods that retain essential attributes—such as posture and scene context—while effectively pseudonymizing individuals by in-place synthesis.

While synthetic data generation has emerged as a potential solution to privacy concerns~\citep{johnson2016mimic, clifton2022roadmap, ridgeway2021challenge, decouple}, generating images entirely from scratch can lead to a data distribution gap and may not retain the valuable context of real-world scenes. 
In pseudonimization, maintaining key attributes—such as pose—is essential while substituting recognizable individuals within the image, a challenge better addressed by in-place generation methods. For example, preserving the original pose is crucial in human-centric images like someone serving in tennis to maintain scene context (see Fig.~\ref{fig:intro_fig}a). Techniques like Stable Diffusion (SD)~\citep{ho20, rombach22} produce realistic images but lack precise control over posture. Conversely, rendering methods~\citep{sariyildiz2023fake} offer fine-grained control over posture but produce less realistic textures, leading to a data distribution gap.

To address these limitations, we propose a novel posture-preserving image pseudonymization pipeline, \textbf{Rendering-Refined Stable Diffusion (RefSD)}. Our key idea is to leverage the strengths of both rendering and diffusion models to create pseudonymized images that retain the original posture and scene context while ensuring privacy compliance. RefSD comprises two modular blocks: a \emph{rendering block} and a prompt-based \emph{generative block}. The rendering block preserves the original pose by generating rendered counterparts of human subjects using extracted 3D meshes, ensuring that posture and spatial context remain intact. In the prompt-based generation block, we utilize SD, guided by text prompts, to synthesize humans. Crucially, we incorporate the rendered poses as conditioning inputs into the SD process during generation. This integration allows SD to generate realistic and high-utility pseudonymized images that accurately preserve the original poses while completely replacing identifiable individuals. It maintains posture accuracy and scene context, provides control over appearance attributes, and generates high-quality, realistic images suitable for downstream vision tasks (Fig.~\ref{fig:intro_fig}b). Moreover, the modular design allows for future enhancements as rendering and diffusion methods improve.

We also introduce \textbf{HumanGenAI}, a framework for systematically evaluating and understanding pseudonymization (for human synthesis) from both qualitative and quantitative perspectives. Currently, there is no clear or standardized way to evaluate pseudonymization techniques, making it challenging to assess their effectiveness comprehensively. We believe it is crucial to evaluate pseudonymization both from human perception and computer vision perspectives, as emphasized in recent works~\cite{patwariperceptanon}. In the qualitative component, we use human perception-based evaluations to assess our RefSD pipeline's ability to generate diverse human features and attributes—many not previously studied—guided by text prompts. 
This includes: (1) evaluating prompt complexity and alignment with attributes like age, gender, ethnicity, and emotion; (2) focusing on accurate representation of individual traits in attribute-level generation; (3) testing sensitivity to subtle variations in fine-grained attribute translation; and (4) analyzing fine-grained and broader features like clothing and occupation in full-body attribute evaluation. 
In the quantitative component, we assess the utility of our pseudonymized images for training downstream tasks such as classification and detection. By integrating both human perception and computational evaluations, HumanGenAI provides a comprehensive framework that emphasizes the importance of combining these perspectives in image pseudonymization evaluation.

The remainder of this paper is structured as follows. Section~\ref{sec:rel_work} reviews related work. Section~\ref{sec:refSD} details the proposed RefSD pipeline, and Section~\ref{sec:human_genai_framework} covers the HumanGenAI framework. Experimental results and insights are presented in Section~\ref{sec:experiments}, and the paper is concluded in Section~\ref{sec:conclusion}.

\section{Related Work}
\label{sec:rel_work}

\textbf{Synthetic Data Generation.} Denoising Diffusion Probabilistic Models (DDPMs)~\cite{ho20} and Latent Diffusion Models (LDMs)~\cite{rombach22} have significantly advanced image generation by reducing computational costs while maintaining high visual fidelity. LDMs, widely adopted due to their open-source availability, have inspired numerous models~\cite{nichol2021glide, ramesh2021zero, saharia22, ramesh2022hierarchical, peebles2023scalable}. Recent studies have leveraged synthetic data from diffusion models to improve image classifiers, finding that augmenting real data with generated data enhances robustness and accuracy in downstream tasks~\cite{he22, li22, bansal23, sariyildiz2023fake}. Rendering-based methods~\cite{wood2021fake, sariyildiz2023fake} generate rendered synthetic images from scratch to improve classifier performance. While effective for creating large-scale synthetic datasets, these methods did not address the need for in-place anonymization of existing real-world images. Generating images entirely from scratch can lead to a data distribution gap~\cite{hennicke2024mind, kollias2022abaw} and may not retain the valuable context of real-world scenes, which is critical for tasks requiring scene understanding.

\begin{figure*}[t]
    \centering
    \begin{tikzpicture}
        \node[inner sep=0pt] (image) at (0,0) {\includegraphics[width=1.0\linewidth]{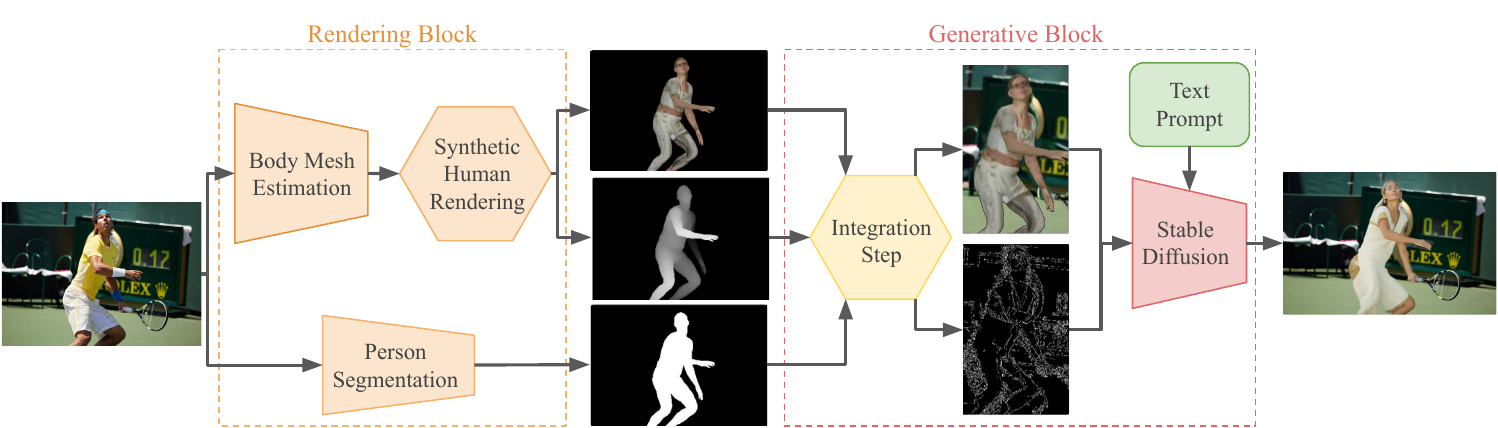}};
        
        \filldraw[fill=yellow, draw=black, thin] (-5.2,1.4) circle (0.17cm);
        \node at (-5.2,1.4) {\small\textbf{1}};

        \filldraw[fill=yellow, draw=black, thin] (-3.2,1.5) circle (0.17cm);
        \node at (-3.2,1.5) {\small\textbf{2}};

        \filldraw[fill=yellow, draw=black, thin] (-4.1,-1) circle (0.17cm);
        \node at (-4.1,-1) {\small\textbf{3}};

        \filldraw[fill=yellow, draw=black, thin] (1.5,-1.3) circle (0.17cm);
        \node at (1.5,-1.3) {\small\textbf{4}};

        \filldraw[fill=yellow, draw=black, thin] (5.1,-1.3) circle (0.17cm);
        \node at (5.1,-1.3) {\small\textbf{5}};

        
    \end{tikzpicture}
    \vspace{-0.6cm} 
    \caption{ Rendering-Refined Stable Diffusion (RefSD) Pipeline: following body mesh estimation \protect\yellowcircle{1}, we render a synthetic human~\citep{pyrender} \protect\yellowcircle{2}. A privacy mask for the original subject is then applied \protect\yellowcircle{3}, merging the synthetic human to replace sensitive data \protect\yellowcircle{4}. Finally, SD generates human-like images with attribute-controlled prompts \protect\yellowcircle{5}.
    }
    \label{fig:pipeline}
    \vspace{-0.5cm}
\end{figure*}

\textbf{Image Pseudonymization.} Traditional pseudonymization techniques like blurring or masking identifiable features~\cite{du2019efficient, yang2022study} are straightforward but often obstruct important visual information, reducing image utility for downstream tasks such as model training~\cite{yang2022study}. Generative models like GANs~\cite{fu2022stylegan, hukkelaas2023realistic, hellmann2024ganonymization} and diffusion models~\cite{klemp2023ldfa, piano2024latent, huang2024humannorm} have advanced realistic human generation but typically focus on generating images from scratch or editing rather than pseudonymizing existing images. Pseudonymization differs from synthetic data generation or image editing; it requires preserving key attributes—such as pose and scene context—while replacing recognizable individuals within the image. The most relevant work to ours is DeepPrivacy2~\cite{hukkelaas2023deepprivacy2}, which pseudonymizes faces and full bodies using separate GAN models—one for faces and another for full bodies—guided by dense pose estimation to retain posture. However, GANs are considered outdated and lack the control and realism of modern diffusion methods. 

Our work bridges this gap by integrating rendering constraints with diffusion models to achieve attribute-guided pseudonymization of human images. This approach ensures precise posture preservation and high-quality image synthesis through in-place human synthesis that maintains the surrounding context. Additionally, it provides the flexibility to modify synthesized human attributes, ensuring compliance without sacrificing image utility.
\section{Rendering-Refined Stable Diffusion~(RefSD)}
\label{sec:refSD}
This section introduces our proposed RefSD pipeline for attribute-guided posture-preserving image pseudonymization for GDPR. We begin with an overview of the RefSD pipeline (Sec.~\ref{subsec:refsd_overview}), followed by detailed descriptions of its key components: the \emph{Rendering Block} (Sec.~\ref{subsec:rendering_block}), \emph{Generative Block} (Sec.~\ref{subsec:generative_block}), and the
integration process (Sec.~\ref{subsec:integration}).

\subsection{Overview}
\label{subsec:refsd_overview}

RefSD is an attribute-guided, posture-preserving image pseudonymization pipeline. It synthesizes human subjects while preserving original posture and scene context. By replacing sensitive data with synthetic representations through in-place generation, RefSD maintains the utility of images for computer vision tasks with minimal contextual disruption. Importantly, it also allows for flexible synthesis of human attributes, facilitating customization of age, ethnicity, and other characteristics as required.

Fig.~\ref{fig:pipeline} shows the pipeline RefSD. It combines a Rendering Block, extracting 3D pose and spatial context, with a Generative Block using Stable Diffusion \cite{ho20, rombach22}. This integration preserves posture and scene context while enabling prompt-based control over pseudonymization, resulting in realistic outputs with flexible attribute modification.

\begin{figure}[t]
    \centering
    \includegraphics[width=\linewidth]{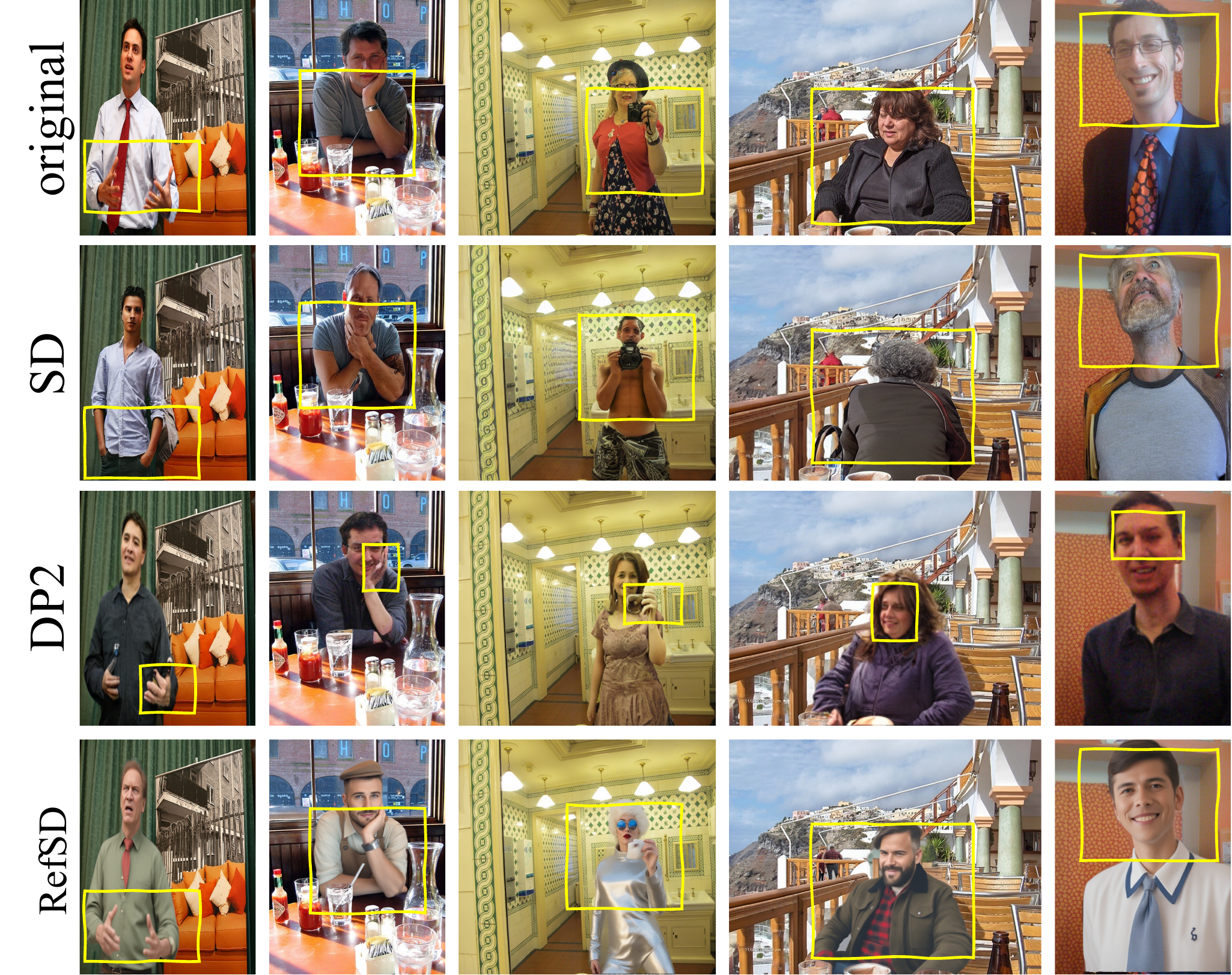}
    \vspace{-0.6cm}
    \caption{Comparison of SD~\cite{rombach2022high}, DP2~\cite{hukkelaas2023deepprivacy2}, and our RefSD for posture-preserving pseudonymization. RefSD achieves superior alignment and realism. More in \texttt{Supplementary Material}.} \label{fig:Posture-preservation}
    \vspace{-0.5cm}
\end{figure}

Unlike existing GAN-based approach, DeepPrivacy2 \cite{hukkelaas2023deepprivacy2}, RefSD uses diffusion models, overcoming limitations in fine-grained realism and lack of attribute control. This results in superior posture fidelity and overall quality compared to both standard SD and DP2. Examples of these methods are provided in Fig.~\ref{fig:Posture-preservation}, where RefSD demonstrates improved performance in preserving whole-body posture, gestures, and facial details.

\subsection{Rendering Block}
\label{subsec:rendering_block}

For each human subject $i$ in image $x$, we extract SMPL parameters $\{\theta_i, \beta_i\}$ using 4DHuman~\citep{goel2023humans}. We generate personalized anonymization masks $a_i$ and bounding boxes $b_i$ using $\mathcal{M}(x, i)$, where $\mathcal{M}$ is a detection and segmentation model~\citep{marwood2023diversity}. Finally, synthetic avatars are rendered using various appearance textures:

\begin{equation} m_i = \mathcal{R}(\theta_i, \beta_i), \quad i = 1, \dots, n \end{equation}

where $m_i$ is the rendered mesh image for subject $i$, preserving posture and shape without identifiable features. We apply a Gaussian filter to the person masks to improve reintegration via alpha blending: $a_i \leftarrow f(a_i)$. 

\subsection{Generative Block}
\label{subsec:generative_block}

We use $\mathcal{G}$ (based on Stable Diffusion XL (SDXL)~\citep{podell2023sdxl}) as our generative model, which can be updated with future SD models~\cite{sehwag2024stretching}. 
For each subject, we extract additional attributes $s_i$ from the SMPL parameters using $\text{PromptAttr}()$, which converts pose and shape information into textual descriptions. We then augment the input prompts with these attributes: $t_i' \leftarrow s_i \oplus t_i$.
We prepare the input for the diffusion model:

\begin{equation} x'_{[\text{crop}_i]} = [x \odot (1 - a_i) + m_i \odot a_i]_{b_i}, \end{equation}

where $[\cdot]_{b_i}$ denotes cropping with bounding box $b_i$. This $x'_{[\text{crop}_i]}$ is scaled to the input size expected by $\mathcal{G}$. To maintain structural fidelity, we generate edge guidance $e_i$ using Canny edge detection on $m_i$, leveraging ControlNet~\cite{zhang2023adding}. The SDXL model $\mathcal{G}$ is then applied:

\begin{equation} \hat{x}_{[\text{crop}_i]} = \mathcal{G}(x'_{[\text{crop}_i]}, e_i, t_i'), \end{equation}

\subsection{Integration Process}
\label{subsec:integration}

We then combine all masks: $a = \bigcup_{i=1}^n a_i$. The final pseudonymized image is reconstructed as:

\begin{equation} \hat{x} = x \odot (1 - a) + \sum_{i=1}^n (\hat{x}_i \odot a_i), \end{equation}

Finally, we detect additional personally identifiable information (PII), represented as $r \leftarrow \mathcal{S}(x, d_{\text{PII}})$, where $\mathcal{S}$ is zero-shot segmentation with Grounding DINO~\citep{liu2023grounding}. These remaining PII areas are then filled with context-matching content using the stable diffusion model and a generic prompt, such that $ \hat{x} \leftarrow \mathcal{A}_{\text{PII}}(\hat{x}, r) $, ensuring they blend seamlessly with the background.

RefSD combines precise posture preservation with customizable, high-quality synthesis, producing GDPR-compliant images that retain essential visual context and utility for downstream vision tasks. The complete pseudocode for the RefSD process is provided in Algo.~\ref{alg:HumanAnonymization}.

\begin{algorithm}[t]
    \DontPrintSemicolon 
    \small
    \caption{Virtual Human Replacement}\label{alg:HumanAnonymization}
    \KwInput{Image $x$ containing $n$ human subjects; Text prompts $\{t_1, \ldots, t_n\}$; PII descriptions $d_{\text{PII}}$}
    \KwResult{Pseudonymized image $\hat{x}$}
    
    \For{$i \gets 1$ \KwTo $n$}{
        $\theta_i, \beta_i \gets \text{4DHuman}(x, i)$  \Comment*{Extract SMPL parameters}
        $a_i, b_i \gets \mathcal{M}(x, i)$  \Comment*{Generate mask and bounding box}
        $a_i \gets f(a_i)$   \Comment*{Mask feathering with gaussian}
        $m_i \gets \mathcal{R}(\theta_i, \beta_i)$  \Comment*{Render synthetic avatar}
    }
    
    \For{$i \gets 1$ \KwTo $n$}{
        $s_i \gets \text{PromptAttr}(\text{SMPL}(\theta_i, \beta_i))$ \\
        $t_i' \gets s_i \oplus t_i$  \Comment*{Extending prompt with orientation}
        $x'_{[\text{crop}_i]} \gets [x \odot (1 - a_i) + m_i \odot a_i]_{b_i}$  \Comment*{Crop with $b_i$}
        $e_i \gets \text{CannyEdge}(m_i)$  \Comment*{Generate edge guidance}
        $\hat{x}_{[\text{crop}_i]} \gets \mathcal{G}(x'_{[\text{crop}_i]}, e_i, t_i')$  \Comment*{Apply diffusion model}
    }
    
    $a \gets \bigcup_{i=1}^n a_i$  \Comment*{Combine masks}
    $\hat{x} \gets x \odot (1 - a) + \sum_{i=1}^n (\hat{x}_i \odot a_i)$  \Comment*{Reintegrate}
    
    $r \gets \mathcal{S}(x, d_{\text{PII}})$  \Comment*{Detect other PIIs}
    $\hat{x} \leftarrow \mathcal{A}_{\text{PII}}(\hat{x}, r)$  \Comment*{Anonymize other PIIs}
    
    \Return $\hat{x}$ 
\end{algorithm}

\section{HumanGenAI Framework}
\label{sec:human_genai_framework}

\begin{figure}
    \centering
    \begin{tikzpicture}
        \node[anchor=south west, inner sep=0] (image) at (0,0) {\includegraphics[width=0.99\linewidth]{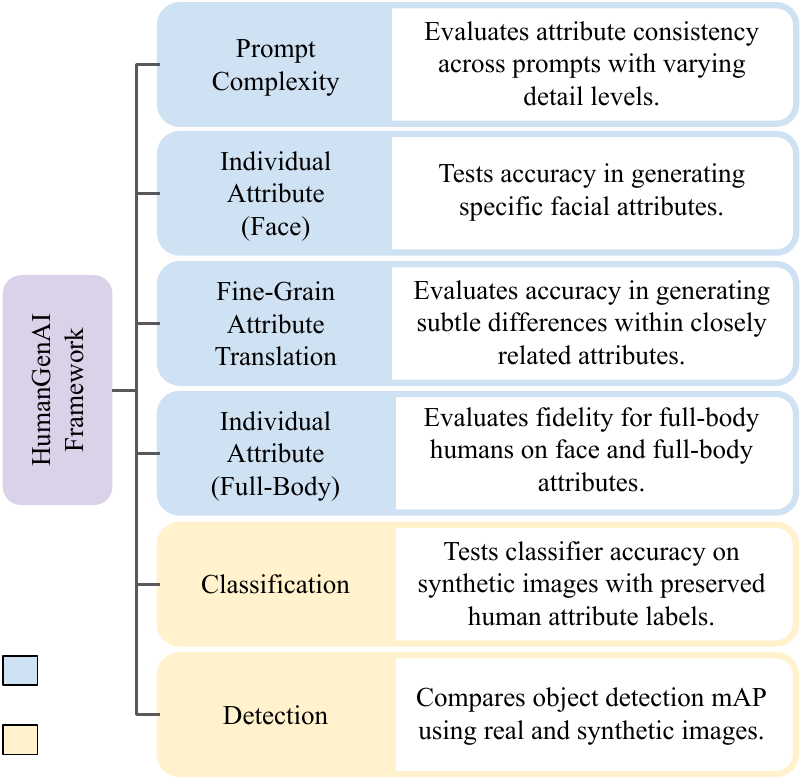}};
        \begin{scope}[x={(image.south east)},y={(image.north west)}]
            \node at (0.23, 0.96) {$\phi_{\text{A}}$}; 
            \node at (0.23, 0.79) {$\phi_{\text{B}}$};
            \node at (0.23, 0.63) {$\phi_{\text{C}}$};
            \node at (0.23, 0.47) {$\phi_{\text{D}}$};
            \node at (0.23, 0.30) {$\psi_{\text{A}}$};
            \node at (0.23, 0.13) {$\psi_{\text{B}}$};
            \node at (0.08, 0.14) {$\phi$};
            \node at (0.08, 0.05) {$\psi$};
        \end{scope}
    \end{tikzpicture}
    \vspace{-0.1cm} 
    \caption{Overview of the HumanGenAI framework. $\phi$: human (annotator) perception evaluations, $\psi$: vision training evaluations.} \label{fig:HumanGenAI_framework}
    \vspace{-0.5cm}
\end{figure}

Evaluating generative models, especially latent diffusion models, is challenging due to difficulties in reliably quantifying consistency, attribute fidelity, and realism~\cite{po2024diffusion, mademlis2024advances, manduchi2024challenges}. The lack of standardized metrics complicates these assessments, often leading to subjective interpretations of quality and fidelity. To address these gaps, we propose HumanGenAI (Fig.~\ref{fig:HumanGenAI_framework}), a framework tailored to evaluate synthetic human generation across key dimensions: \textbf{human attribute fidelity} (Sec.~\ref{subsec:human_perception_evals}) and \textbf{image utility for downstream tasks} (Sec.~\ref{subsec:utility_evals}). This structured approach enables comprehensive assessment aligned with the diverse specific requirements of human image generation. Evaluation details are in Sec.~\ref{subsec:humangenai_details}.

\subsection{Human Perception Evaluations}
\label{subsec:human_perception_evals}

Recent works~\citep{patwariperceptanon, sun2024privacy} highlight the critical role of human (annotator) evaluations in assessing image privacy and synthetic human generation, emphasizing the importance of aligning generated images with human expectations. Given this, we design four distinct experiments to evaluate different aspects of human attribute fidelity in generated images, relying on human annotators for assessments.

\textbf{Prompt Complexity ($\phi_{\text{A}}$).} 
This evaluation examines how the detail level of prompts (\textit{simple, medium, complex}) affects generated face images. Using identical source images, we compare results across prompt types, each varying in descriptive richness, to assess consistency in representing a combination of key attributes—age, ethnicity, gender, emotion, and face attributes. Annotators score each image on how well it aligns with the intended attributes.

\textbf{Individual Attribute -- Face ($\phi_{\text{B}}$).} 
This test focuses on the precision in generating individual facial attributes, isolating each attribute type (\textit{emotion, ethnicity, or face Characteristics}) in the prompt to evaluate accuracy. A set of 50 specific attributes is tested, where annotators assess how each attribute is represented in the generated images.

\textbf{Fine-grained Attribute Translation ($\phi_{\text{C}}$).}
This evaluation aims to study subtle variations within specific attribute categories: age, ethnicity, emotion, and skin tone. Using closely related attribute pairs (e.g., \textit{similar ethnicities or adjacent age groups}), we measure the RefSD's capability to generate fine distinctions. Side-by-side generated images are presented to annotators to verify attribute fidelity.

\textbf{Individual Attribute -- Full-Body ($\phi_{\text{D}}$).}
Extending from $\phi_{\text{B}}$, this evaluation considers a broader array of full-body characteristics. Attributes include gender, age, emotion, ethnicity, skin tone, and additional categories like clothing style and occupation, totaling 100 subcategories. Annotators assess each generated full-body image for alignment with specified attributes, allowing a comprehensive review of attribute versatility and granularity in synthesis.

The details of the attribute list for these four experiments can be found in the \texttt{Supplementary Material}. These evaluations ensure that the synthetic outputs are not only technically accurate but also perceptually aligned with intended attribute representations.

\subsection{Utility Evaluations}
\label{subsec:utility_evals}

Following standard utility experiments, our evaluations assess the impact of pseudonymized RefSD-generated data on downstream classification and detection tasks. Using existing datasets, we replace real images with synthetic counterparts and compare model training performance between pseudonymized and original data to evaluate efficacy.

\textbf{Utility Training: Classification} ($\psi_{\text{A}}$).
We generate synthetic images using prompts that incorporate original dataset labels to preserve key attributes for classifier training. Classifier performance is then evaluated on a real-image test set to assess the effectiveness of synthetic data for attribute-specific model training. We use RAF-DB~\cite{rafdb}, which provides labeled images for Age, Emotion, Ethnicity, and Gender, to build classifiers for each attribute.

\textbf{Utility Training: Detection} ($\psi_{\text{B}}$). We synthesize pseudonymized images by replacing all human subjects, then train multi-object detectors on this synthetic data and test on real images to evaluate detection performance. For this task, we use the OpenImages~\cite{openimages} dataset, comparing results with models trained on real-world data to assess the effectiveness of training detectors on our human pseudonymized images.

\subsection{HumanGenAI Details}
\label{subsec:humangenai_details}

\textbf{Image Collection.}
Our HumanGenAI framework uses a curated set of source images from CelebA~\citep{celeba}, RAF-DB~\citep{rafdb}, Chicago Face Dataset (CDF)~\cite{ma2015chicago}, and Flickr-Faces-HQ (FFHQ)~\cite{karras2019style} for face images, and COCO~\citep{mscoco}, VOC~\citep{voc}, and OpenImages~\citep{openimages} for full-body images. Face datasets provide frontal views with detailed attribute labels, while full-body datasets contain multi-object scenes with humans. These datasets support varied attribute specificity across different body regions.

\textbf{Prompt Design.}
To drive image generation, we designed four prompt templates—basic, simple, medium, and complex—each incorporating varying degrees of attribute detail. The basic prompt structure, e.g., “\texttt{A White person}” or “\texttt{A person with a goatee},” specifies a single attribute. In contrast, the simple prompt includes five attributes, such as “\texttt{A 95-year-old White Female with brown hair, showing an Angry emotion}.” Medium and complex prompts expand on this structure, adding qualifiers to enhance clarity and fidelity. Further prompt templates and examples can be found in the \texttt{Supplementary Material}.

\textbf{Human Annotations.}
Each generated image was rated by three annotators on a 1-5 scale to assess alignment with specified attributes, consistent with standards in image privacy assessment~\citep{sun2024privacy, patwariperceptanon}. Unlike binary ratings, a graded scale better captures human perception nuances, as shown in recent anonymization studies~\citep{patwariperceptanon}. While previous work used a 10-point scale, our preliminary testing showed that 5 points effectively reflect annotators' preferences for attribute satisfaction and prompt alignment.
\section{Experiments}
\label{sec:experiments}

In this section, we present the experimental setup, followed by a detailed analysis of each HumanGenAI evaluation scenario: human perception ($\phi$) and utility ($\psi$) evaluations, as outlined in Section~\ref{sec:human_genai_framework}. Complete set of figures are provided in high resolution in the \texttt{Supplementary Material}. We conclude with discussions on key aspects of RefSD.

\begin{table}[t] \scriptsize
\centering
 \caption{HumanGenAI experimental details. Prompt types: Basic (B), Simple (S), Medium (M), and Complex (C). Datasets: CelebA (C), COCO (CO), Pascal VOC (P), RAF-DB (R), OpenImages (O), CDF (CD), FFHQ (F).}
\label{tab:dataset_details}
\vspace{-0.3cm}
\setlength{\tabcolsep}{1.4mm}{

\begin{tabular}{l|cccccc|c}
\toprule
          & $\phi_{\text{A}}$ & $\phi_{\text{B}}$ & $\phi_{\text{C}}$ & $\phi_{\text{D}}$ & $\psi_{\text{A}}$ & $\psi_{\text{B}}$ & Total\\

\midrule
\# Src Imgs & 33 & 250 & 250 & 250 & 12,271 & 11,545 & $-$\\
Type  & Faces  & Faces  & Faces  & Full & Full & Full & $-$\\
Src Dataset & C  & C,CD,F  & C,CD,F  & CO,P & R & O & $-$\\
\midrule
\# Prompts & 1,188 & 50 & 33 & 100 & 12,271 & 11,545 & $-$ \\
Type & S,M,C & B & S & S & S & S & $-$ \\
\midrule
\# Syn. Imgs & 3,564$^2$ & 12,500 & 8,250 & 25,000 & 12,271 & 11,545 & \textbf{73,130}\\
Annotators & 3 &  3 & 3 & 3 & $-$ & $-$ & \textbf{8}$^3$\\
All Anno & 10,692 & 37,500 & 24,750 & 75,000 & $-$ & $-$ &\textbf{147,942}\\
\bottomrule
\end{tabular}}
\vspace{-0.5cm}
\end{table}

\textbf{Experimental Setup.}
Table~\ref{tab:dataset_details} summarizes datasets, generated images, and annotations. For human perception evaluations ($\phi$), annotators scored image alignment on a 1-5 scale (5 being best), with reliability assessed via Cronbach’s Alpha~\citep{tavakol2011making}.
For classification ($\psi_{\text{A}}$), ViT-tiny and ViT-base models were trained on synthetic, real, combined, and pretrain/finetune (synthetic-real) configurations using RAF-DB’s train set (12,271 images), with ground-truth emotion, ethnicity, gender, and age labels embedded in RefSD prompts. Evaluation was performed on the RAF-DB test set (3,068 images) using accuracy.
For detection ($\psi_{\text{B}}$), we trained object detectors—DINOv2-Adapter~\cite{oquab2023dinov2} encoder and Faster RCNN~\cite{ren2016faster}—on synthetic, real, and pretrain/finetune (synthetic-real) settings using a subset of OpenImages (75,000 images) with 11,545 synthesized or blurred human/license plate instances. The model was evaluated on 1,564 validation images covering 600 classes, including 227 Human Faces and 722 Persons. The 73,130 image generation took 5 days using 4$\times$H100 GPUs. More details are provided in the \texttt{Supplementary Material}.

\footnote[]{$^{2}$33 source images with 1,188 prompts, totaling 3,564 samples.}
\footnote[]{$^{3}$ $\phi_{\text{A}}$ to $\phi_{\text{D}}$ each: 3 annotators randomly chosen from total pool of 8.}

\subsection{Human Perception Evaluations}

\begin{figure}[t]
    \centering
    \includegraphics[width=0.9\linewidth]{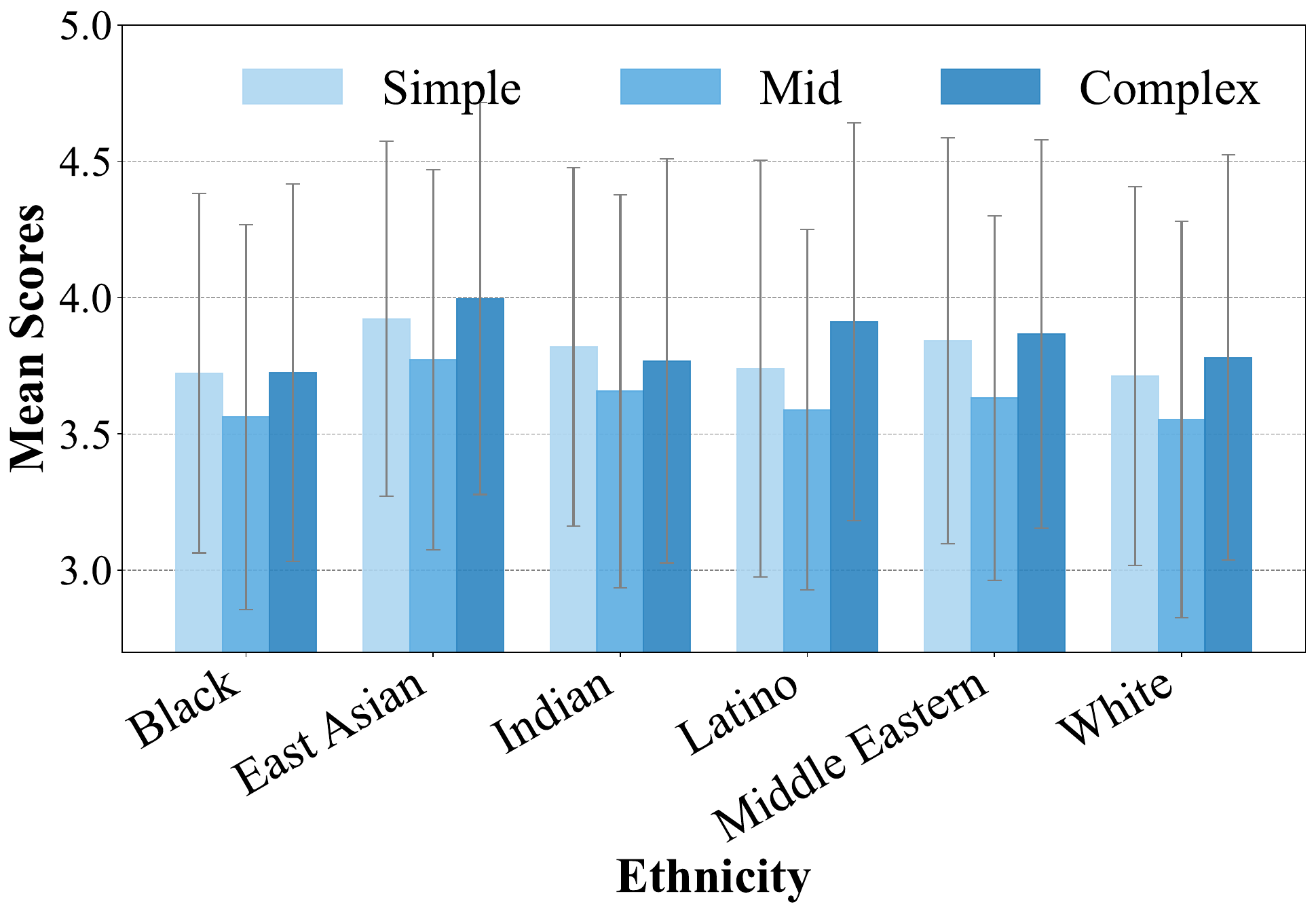}
    \vspace{-0.3cm}
    \caption{Mean annotator scores for \textbf{Prompt Complexity ($\phi_{\text{A}}$)} for Ethnicity. Remaining (Age, Gender, Emotion, Face Attributes) provided in \texttt{Supplementary Material}.}
    \label{fig:phi_1_ethnicity}
    \vspace{-0.6cm}
\end{figure}

\textbf{Prompt Complexity ($\phi_{\text{A}}$).}
Fig.~\ref{fig:phi_1_ethnicity} shows that complex prompts slightly outperform simple prompts in mean annotator scores, though the difference is minor. Simple prompts sometimes yield the highest attribute accuracy, while medium prompts consistently score lowest. The similar performance of simple and complex prompts suggests that added detail does not significantly enhance image accuracy or quality. Annotator consistency scores were 0.773 (simple), 0.728 (medium), and 0.727 (complex).

\noindent \emph{Insights:} These results suggest that prompt complexity minimally affects RefSD's attribute generation, indicating a possible ceiling in its ability to interpret nuanced prompts. Interestingly, medium prompts are less effective than simple or complex ones—perhaps because complex prompts offer stronger guidance, while simple prompts are straightforward to interpret accurately.

\begin{figure*}[t]
    \centering
    \includegraphics[width=0.88\linewidth]{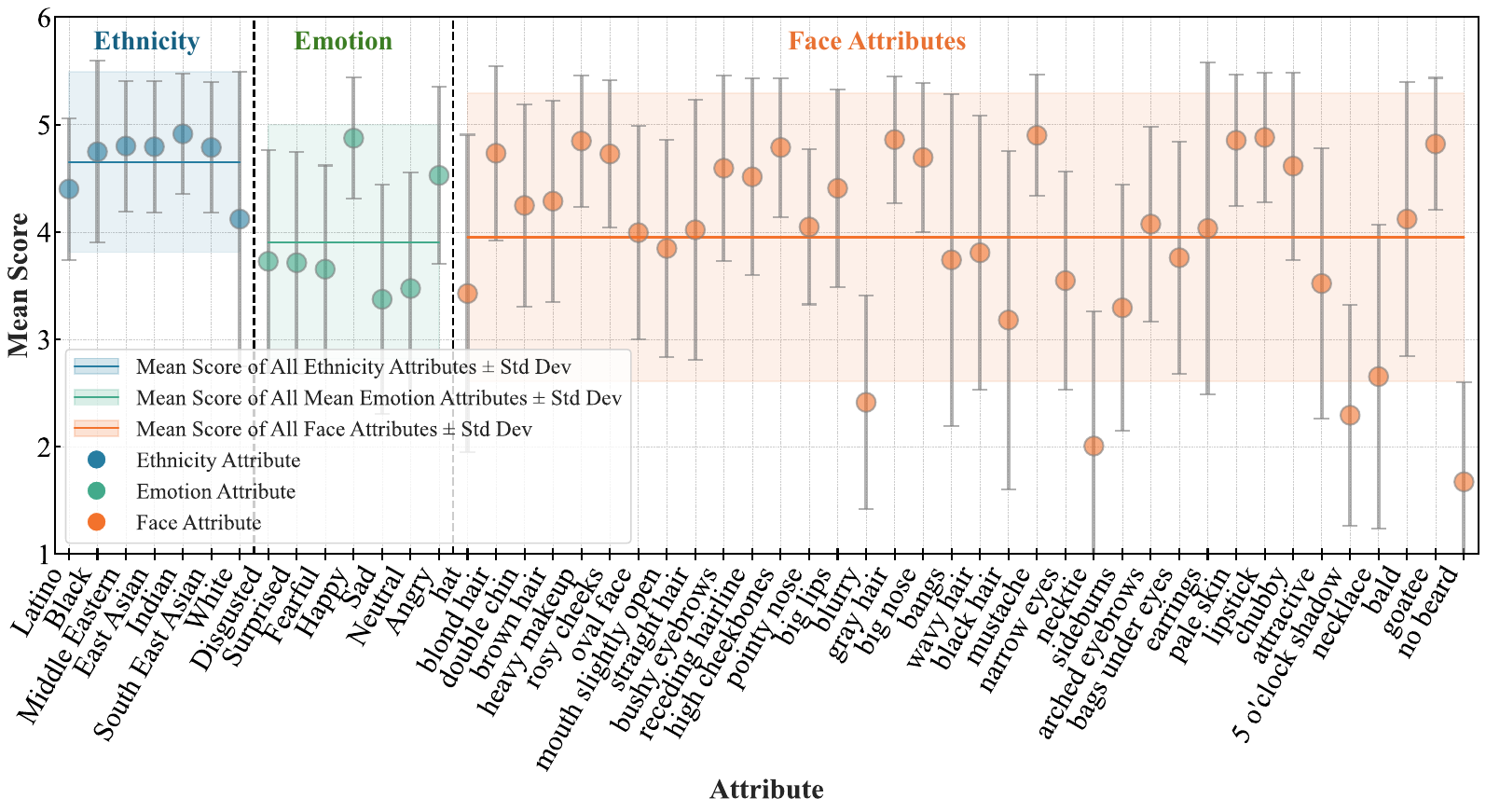}
    \vspace{-0.3cm}
    \caption{Mean scores of annotations for \textbf{Individual Attribute -- Face ($\phi_{\text{B}}$)} split by categories: ethnicity, emotion, and face attributes. We display the average (represented by color lines) of all attributes within each category.} \label{fig:phi_2}
    \vspace{-0.6cm}
\end{figure*}

\begin{figure}[h]
    \centering
    \includegraphics[width=1.0\linewidth]{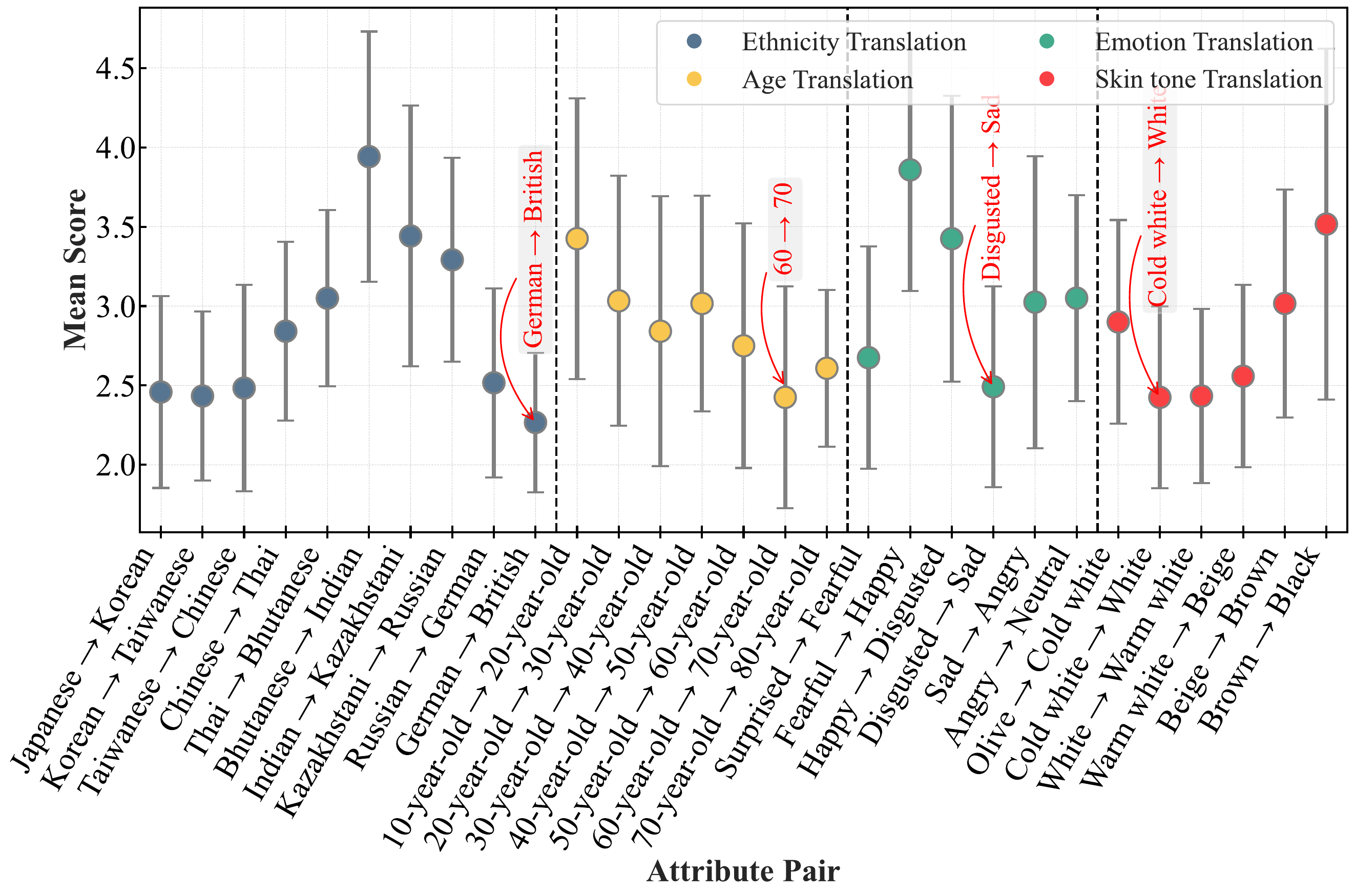}
    \vspace{-0.6cm}
    \caption{Mean annotation scores for \textbf{Fine-Grained Attribute Translation ($\phi_{\text{C}}$)} across Ethnicity, Emotion, Age, and Skin tone groups. Highlighted are the pairs with the lowest mean scores.} \label{fig:phi_3}
    \vspace{-0.6cm}
\end{figure}

\textbf{Individual Attribute -- Face ($\phi_{\text{B}}$).}
Fig.~\ref{fig:phi_2} shows mean scores for ethnicity, emotion, and facial features. Ethnicity attributes scored consistently above 4.5, indicating effective representation. While distinct emotions (e.g., angry, happy) were well-captured, subtler ones showed lower agreement, possibly due to limited data coverage. Facial attributes had mixed results: prominent features (e.g., goatee, blond hair) scored well, while niche attributes (e.g., necktie, 5 o’clock shadow) were less reliable. Prompts for no beard often generated beards, suggesting a need for more explicit cues. Overall annotator consistency was 0.752.

\noindent \emph{Insights:} These findings indicate that RefSD effectively generates well-defined emotions and ethnic diversity. However, like other image generation methods, it struggles with subtle emotions and specific facial features because certain ambiguous attributes (e.g., surprise, eye types) are hard to differentiate. Its higher effectiveness in generating ethnicity attributes may be because ethnicities are more prominently represented and easier to identify. Therefore, it is important to consider attribute complexity, such as complexity of emotions that may occur simultaneously.

\textbf{Fine-grain Attribute Translation ($\phi_{\text{C}}$).}
This experiment assesses RefSD's ability to capture subtle variations across ethnicity, age, emotion, and skin tone. Fig.~\ref{fig:phi_3} shows mean scores, revealing challenges in distinguishing closely related ethnicities (e.g., Japanese vs. Korean, German vs. English) and decreased sensitivity to aging cues in older groups. In contrast, the model performed better with contrasting skin tones, improving progressively from lighter to darker shades. Annotator consistency was 0.340, indicating limited reliability in subtle attribute differentiation.

\noindent \emph{Insights:} The HumanGenAI evaluation framework effectively identifies challenging attribute pairs in fine-grained translation for RefSD, particularly those with high correlation.
For example, for similar skin tones, such as \textit{Cold White and White}, and differentiating similar ethnicities, the model often generates nearly identical images. This showcases areas for improvement in Stable Diffusion models, emphasizing the need for more nuanced data representation.

\begin{figure}[t]
    \centering
\includegraphics[width=1.0\linewidth]{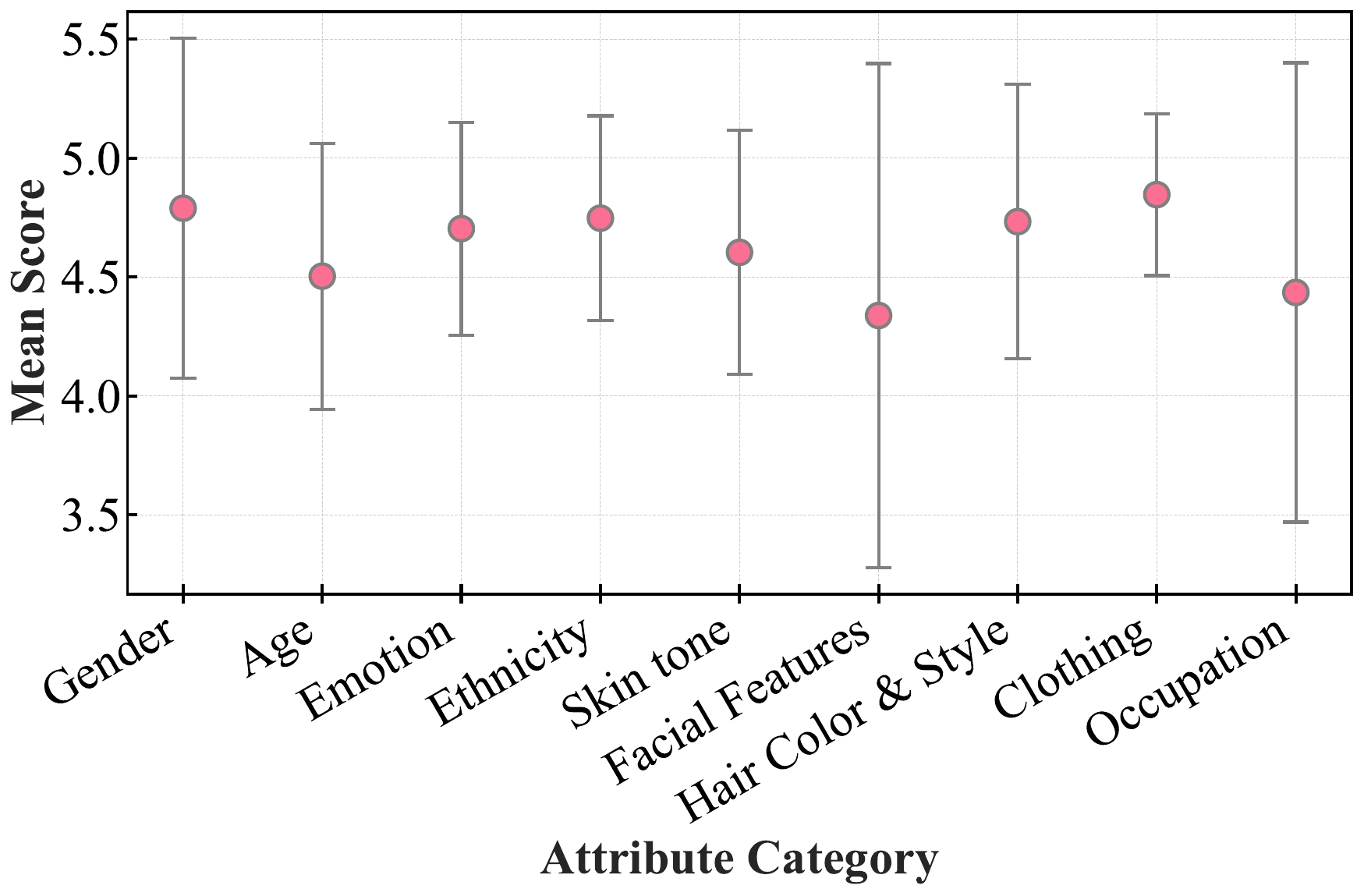}
    \vspace{-0.6cm}
    \caption{Annotator mean scores for \textbf{Individual Attribute -- Full-Body ($\phi_{\text{D}}$)}, grouped based on the meta-attribute categories. Individual figures are provided in \texttt{Supplementary Material}.} \label{fig:phi_4}
    \vspace{-0.6cm}
\end{figure}

\textbf{Individual Attribute -- Full-Body ($\phi_{\text{D}}$).} Figure~\ref{fig:phi_4} shows the mean annotator scores across attribute categories for full-body images. Overall scores are high; emotion, ethnicity, and clothing achieve high means and low standard deviations, indicating consistent generation quality. In contrast, facial features and occupation exhibit larger standard deviations and lower mean scores, reflecting variability in generation accuracy or satisfaction. Occupations with distinctive uniforms (e.g., clown, firefighter) receive higher scores, while those without unique attire (e.g., butcher, bartender) score lower. This suggests RefSD relies on strong visual identifiers for accurate representations.

\noindent \emph{Insights:} RefSD effectively captures and reproduces attributes with clear and distinctive visual cues, such as clothing, resulting in high consistency and accuracy. However, facial features are harder to identify and implement in full-body images because faces are smaller and require more fine-grained control. This leads to greater variability in attributes that need subtle or detailed representations.

\subsection{Utility Evaluations}

\textbf{Utility Training: Classification ($\psi_{\text{A}}$).} Table~\ref{tab:rafdb1} shows classification accuracies on the RAF-DB test set for Emotion, Age, Gender, and Ethnicity using ViT-Tiny and ViT-Base models. Training methods include synthetic (S), real (R), synthetic pre-training then real fine-tuning (S$\rightarrow$R), and combined training (S$+$R). Incorporating synthetic data consistently enhances performance across all attributes. Notably, pre-training on synthetic and fine-tuning on real or combining both lead to accuracy improvements ranging from 0.5\% to 11.1\%. For instance, ViT-Base shows up to a 3.8\% increase in Emotion and an 11.1\% boost in Gender classification compared to training solely on real data.

\noindent \emph{Insights:} \textbf{Leveraging synthetic data from RefSD consistently improves attribute classification accuracy.} Pre-training with synthetic data followed by real data fine-tuning offers the most substantial gains, highlighting the complementary role of synthetic and real datasets in enhancing model robustness and performance.

\begin{table}[t]\scriptsize
\caption{\textbf{Utility Training: Classification ($\psi_{\text{A}}$).} Classification accuracy (\%) on the RAF-DB test set. Classifiers trained on RefSD pseudonymized synthetic (S) vs. real-world (R) data. S$\rightarrow$R: pre-training on synthetic, fine-tuning on real; S$+$R: mixed training.
    \label{tab:rafdb1}}
\vspace{-0.6cm} 
\centering
\setlength{\tabcolsep}{2.2mm}{
\begin{tabular}{@{}lcccccccc@{}}
\toprule
\multirow{2}{*}{Model}   & \multicolumn{4}{c}{Emotion} & \multicolumn{4}{c}{Age} \\
\cmidrule(lr){2-5} \cmidrule(lr){6-9}
          & S & R & \makecell[c]{S$\rightarrow$R} &  \makecell[c]{S$+$R}& S & R & \makecell[c]{S$\rightarrow$R} &  \makecell[c]{S$+$R}\\
\midrule
ViT-Tiny & 39.6 & 41.5 & \textbf{42.2} & 42.0 & 48.4 & 57.0 & 55.7 & \textbf{58.5} \\
ViT-Base & 36.3 & 41.5 & \textbf{45.3} & 44.3 & 48.2 & 58.4 & 58.1 & \textbf{59.9} \\
\bottomrule
\toprule
\multirow{2}{*}{Model}  & \multicolumn{4}{c}{Gender} & \multicolumn{4}{c}{Ethnicity} \\
\cmidrule(lr){2-5} \cmidrule(lr){6-9}
          & S & R & \makecell[c]{S$\rightarrow$R} &  \makecell[c]{S$+$R}& S & R & \makecell[c]{S$\rightarrow$R} &  \makecell[c]{S$+$R}\\
\midrule
ViT-Tiny & 52.9 & 60.6 & \textbf{65.1} & 63.4 & 68.2 & 77.5 & \textbf{77.6} & 77.5 \\
ViT-Base & 53.1 & 61.9 & 64.4 & \textbf{73.0} & 67.6 & 78.2 & 78.8 & \textbf{79.9} \\
\bottomrule
\end{tabular}}
\vspace{-0.1cm}
\end{table}

\textbf{Utility Training: Detection ($\psi_{\text{B}}$).} 
Our results in Table~\ref{tab:map_results} demonstrate that detectors trained on RefSD pseudonymized images achieved consistently higher mAP scores than those trained on real images, with a 1.1\% gain in mAP@[.5:.95] and a 1\% gain in mAP@0.5. This indicates that RefSD images provide not only privacy benefits but also competitive utility. When RefSD data was used for pretraining before fine-tuning on real data, mAP further improved by 5.5\% (mAP@[.5:.95]) and 6.6\% (mAP@0.5) compared to training on real data alone. 

\noindent \emph{Insights:} \textbf{RefSD synthetic data can effectively augment datasets, enhancing model performance even on consented data.} Using pseudonymized images consistently improves results without any negative impact, highlighting RefSD’s potential for both privacy and utility.

\begin{table}[t] \footnotesize
\caption{\textbf{Utility Training: Detection ($\psi_{\text{B}})$.} Mean Average Precision (mAP) comparison between pseudonymized synthetic and real-world data (OpenImages). $\rightarrow$ denotes pre-training on synthetic, followed by fine-tuning on real data.}
\label{tab:map_results}
\vspace{-0.3cm} 
\centering
\setlength{\tabcolsep}{7mm}{
\begin{tabular}{@{}lccc@{}}
\toprule
Metric             & S & R & S $\rightarrow$ R  \\
\midrule
mAP@[.5 : .95] ↑     & 26.4 & 25.3 & \textbf{30.8} \\
mAP@0.5 ↑          & 33.2 & 32.2 & \textbf{38.8} \\
\bottomrule
\end{tabular}}
\vspace{-0.2cm} 
\end{table}

\subsection{Discussion}
\label{subsec:discussion}

This section discusses critical dimensions in synthetic human generation using RefSD, focusing on bias, diversity, prompt control, and privacy risks. These aspects are fundamental to achieving privacy-compliant, high-utility pseudonymization while considering the ethical and practical limitations of generative models. 

\noindent \emph{\textendash} \textbf{Bias \& diversity.} Unconstrained prompt-based generation can lead to biases and repetitive patterns inherent in diffusion models~\citep{marwood2023diversity}. To counter this, RefSD relies on pseudonymizing existing images as latent space encodings, providing a diverse source that avoids SD model local minima. By integrating rendered meshes and personalized prompts, we introduce variability and enhance control and diversity in generated attributes.

\noindent \emph{\textendash} \textbf{Inheriting SD's bias \& fairness issues.}
Our RefSD pipeline is modular, with separate rendering and generation blocks, currently using SDXL for generation. This allows easy replacement of SDXL with future models as bias and fairness research advances. While RefSD may inherit SDXL’s biases, our main objective is to evaluate fine-grained human attributes with current models, using the HumanGenAI framework to identify limitations. Though testing all SD variants is beyond our scope, RefSD provides a flexible pipeline, adaptable to fairer models as they emerge.

\noindent \emph{\textendash} \textbf{Prompt-controlled pseudonymization.}
RefSD offers prompt-controlled pseudonymization, to shape generated human attributes for diversity, context alignment, and label retention, supporting three main strategies: \textit{(1) Random Prompts}: Replaces human subjects with random attributes. \textit{(2) Data Diversification}: Introduces varied and balanced human representations, enhancing dataset diversity. \textit{(3) Attribute Preservation}: Incorporates original labels into prompts to pseudonymize labeled datasets. 
This flexibility allows RefSD to address diverse pseudonymization requirements while preserving data utility and ensuring compliance with GDPR.

\noindent \emph{\textendash} \textbf{Re-identification risks via pose and location.}
Pose and location may present re-identification risks for those familiar with the subject or scene. To balance privacy and utility, we adhere to GDPR guidelines, where fully pseudonymizing pose and location could overly reduce utility. Our approach preserves critical data while acknowledging privacy trade-offs. Recent studies show that background cues like location can add to privacy risks~\cite{patwariperceptanon}, an area we consider for future privacy-aware generation.
\section{Conclusion}
\label{sec:conclusion}

We introduced Rendering-Refined Stable Diffusion, a novel pipeline for in-place human pseudonymization of images by combining 3D-rendered poses with prompt-based latent diffusion. RefSD allows precise manipulation of human attributes—such as age, ethnicity, and emotion—during pseudonymization while preserving the original pose and scene context, addressing the limitations of traditional methods that degrade image quality, obscure critical context, or lack synthesis control in human-centric images. To evaluate the effectiveness of attribute customization and utility, we proposed HumanGenAI, a framework that studies human attribute fidelity from a human perception perspective and includes utility experiments for training vision models. Our assessments demonstrate that RefSD accurately synthesizes a variety of unique human attributes guided by text prompts. Our utility results show that combining synthetic data with real data improves classification performance, while synthetic data alone boosts detection accuracy, with greater gains when used together. 

\paragraph{Broader Impact.} 
\label{broader_imp}
We aim for our research to drive advancements in human image generation technologies, fostering developments that are both ethically sound and socially beneficial. Our RefSD pipeline, with its ability to generate and modify human images through prompts, opens up numerous possibilities in creative industries, personalized digital media, and privacy-preservation. However, this powerful technology must be used responsibly to avoid misuse, such as the creation of misleading or harmful content. We emphasize the importance of adhering to ethical guidelines and encourage the development of robust mechanisms to detect and prevent misuse. Additionally, our research highlights the need for continued discourse on the implications of human image generation, including considerations of bias, fairness, and the potential societal impacts. By promoting transparency and ethical standards, we hope to contribute to the responsible advancement of this field.
{
    \small
    \bibliographystyle{ieeenat_fullname}
    \bibliography{main}
}

\clearpage
\maketitlesupplementary

\definecolor{AgeColor}{HTML}{D72638}       
\definecolor{EthnicityColor}{HTML}{3B7DD8} 
\definecolor{GenderColor}{HTML}{FF9F1C}    
\definecolor{FaceAttrColor}{HTML}{008744}  
\definecolor{EmotionColor}{HTML}{9A37FF}   

Rendering-Refined Stable Diffusion (RefSD) is an image pseudonymization pipeline that synthesizes human figures while inpainting other personal identifiable information (PII). The pipeline replaces humans in the original image with 3D-rendered avatars and utilizes Stable Diffusion, constrained by these rendered avatars, to produce realistic synthetic humans.

We first discuss the prompts, their types, and templates (Sec.~\ref{appendix:prompts}), followed by examples of RefSD-generated images and comparative results (Sec.~\ref{appendix:example_imgs}). We provide details on the attribute categories and include the remaining plots for human perception ($\phi$) in Sec.~\ref{appendix:phi}. Some plots are re-presented in high resolution for better visual clarity. We provide more details on utility evaluation ($\psi$) in Sec.~\ref{appendix:psi}, which also contains training details for all models. 


\section{Prompt Templates and Details}
\label{appendix:prompts}

To generate images using Stable Diffusion, we designed four prompt types, each varying in complexity and additional details. Every prompt follows a consistent structure:

\begin{center} \texttt{Prefix + Attribute Prompt + Suffix} \end{center}

\noindent All prompts included a common \textbf{prefix:} \texttt{seen from front,} which was empirically tested to ensure image quality. 

Additionally, to enhance image quality and realism, we used a negative prompt for all images. The common negative prompt used was:

\textbf{Negative prompt:} \textit{`drawing, painting, blurry, smooth, cgi, anime, rendering, black and white, oily, wet, shining light, hard light, special effect, nudity, sexy, erotic, topless, sports clothing''}. 

Each prompt type is described below with examples.

\subsection{Basic Prompt}
The basic prompt contains only one attribute, \texttt{{\color{AgeColor}X}}.
\begin{description}[leftmargin=20pt, labelindent=20pt]
    \item[Template] – \texttt{A {\color{AgeColor}X} person} or  \texttt{A person with {\color{AgeColor}X}}
    \item[Example \#1] – \textit{``A person with {\color{AgeColor}no beard}''}
    \item[Example \#2] – \textit{``A {\color{AgeColor}sad} person.''}
\end{description}

\subsection{Simple Prompt}
The simple prompt contains attributes from 5 categoties; age, ethnicity, gender, face attribute (face attr), and emotion without any additional suffix or details.
\begin{description}[leftmargin=20pt, labelindent=20pt]
    \item[Template] – \texttt{A {\{\color{AgeColor}age}\} {\{\color{EthnicityColor}ethnicity}\} {\{\color{GenderColor}gender}\} with {\{\color{FaceAttrColor}face attr}\}, showing {\{\color{EmotionColor}emotion}\} emotion.}
    \item[Example] – \textit{``A {\color{AgeColor}10-year-old} {\color{EthnicityColor}Indian} {\color{GenderColor}Female} with {\color{FaceAttrColor}rosy cheeks}, showing {\color{EmotionColor}Happy} emotion.''}
    
    
\end{description}

\subsection{Medium Prompt}
The medium prompt contains the same five categories as the simple prompt, but with additional suffixed details to enhance the realism of the image. There is additional emphasis on emotion, which during initial testing was least apparent in generated images.
\begin{description}[leftmargin=20pt, labelindent=20pt]
    \item[Template] – \texttt{A {\{\color{AgeColor}age}\} {\{\color{EthnicityColor}ethnicity}\} {\{\color{GenderColor}gender}\} with {\{\color{FaceAttrColor}face attr}\}, showing a clearly exaggerated {\{\color{EmotionColor}emotion}\} emotion. The portrait is natural and realistic, with sharp focus and high detail.}
    \item[Example] – \textit{``A {\color{AgeColor}10-year-old} {\color{EthnicityColor}Indian} {\color{GenderColor}Female} with {\color{FaceAttrColor}rosy cheeks}, showing a clearly exaggerated {\color{EmotionColor}Happy} emotion. The portrait is natural and realistic, with sharp focus and high detail.''}
\end{description}

\subsection{Complex Prompt}
The complex prompt follows the medium prompt, but expands the additional details to enhance image quality. It has a greater emphasis on emotion.
\begin{description}[leftmargin=20pt, labelindent=20pt]
    \item[Template] – \texttt{A {\{\color{AgeColor}age}\} {\{\color{EthnicityColor}ethnicity}\} {\{\color{GenderColor}gender}\} with {\{\color{FaceAttrColor}face attr}\}, and their face is expressing very exaggerated {\{\color{EmotionColor}emotion}\} emotion. The image is natural, realistic, sharp focus, high detail,  medium format photograph, person , (Nikon DSLR Camera, 8K resolution, Detailed face features).}
    
    \item[Example] – \textit{``A {\color{AgeColor}10-year-old} {\color{EthnicityColor}Indian} {\color{GenderColor}Female} with {\color{FaceAttrColor}rosy cheeks}, and their face is expressing very exaggerated {\color{EmotionColor}Happy} emotion. The image is natural, realistic, sharp focus, high detail,  medium format photograph, person , (Nikon DSLR Camera, 8K resolution, Detailed face features).''}
\end{description}


\section{Comparisons with Related Works}
\label{appendix:example_imgs}

\begin{figure*}[t]
    \centering
    \includegraphics[width=0.9\linewidth]{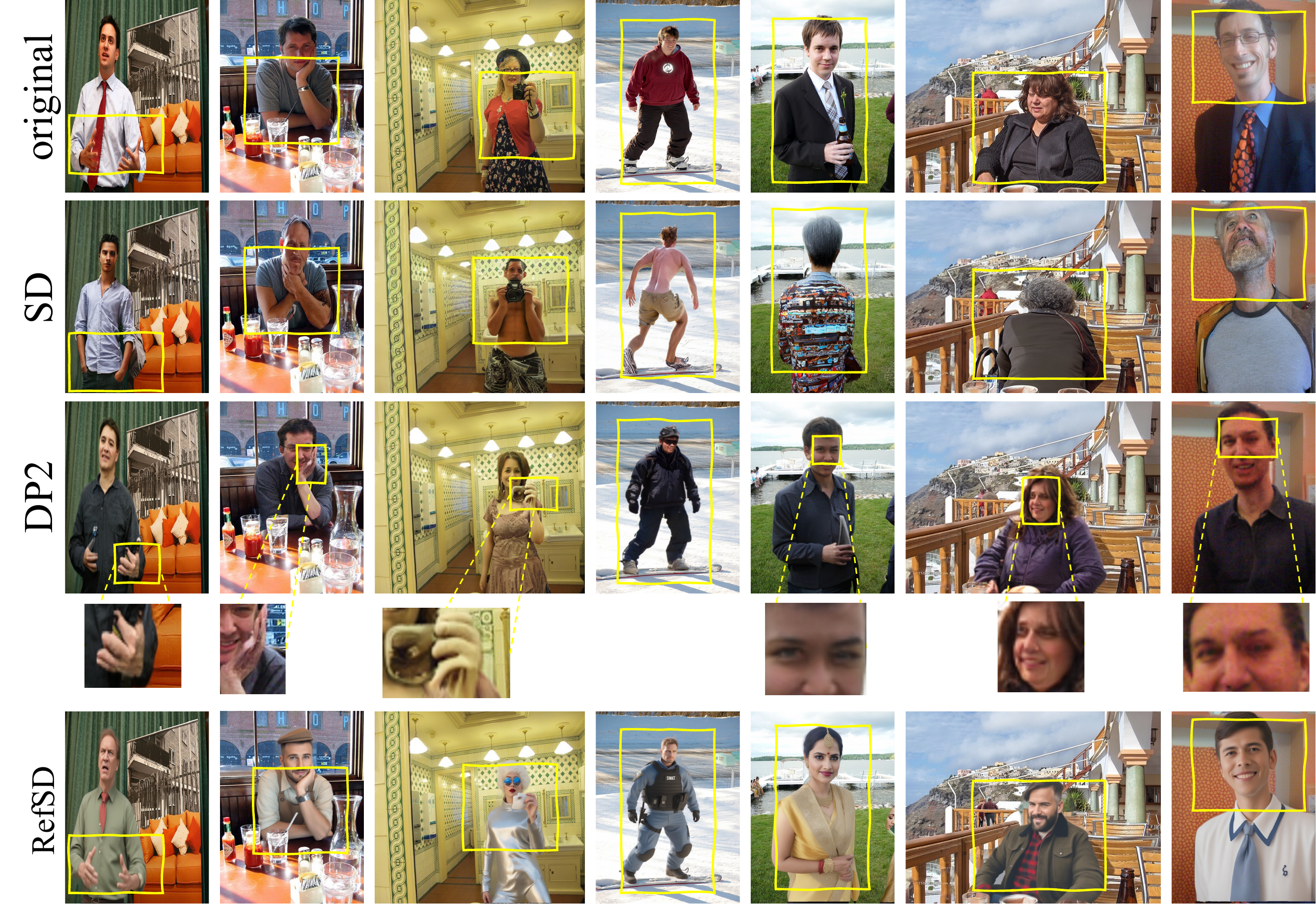}
    \caption{Comparison of regular Stable Diffusion (SD), DeepPrivacy2 (DP2), and our RefSD for posture-preserving pseudonymization. RefSD achieves superior alignment and realism}
    \label{fig:appendix_refsd_comparison}
\end{figure*}

\begin{figure}[h]
    \centering
    \includegraphics[width=1\linewidth]{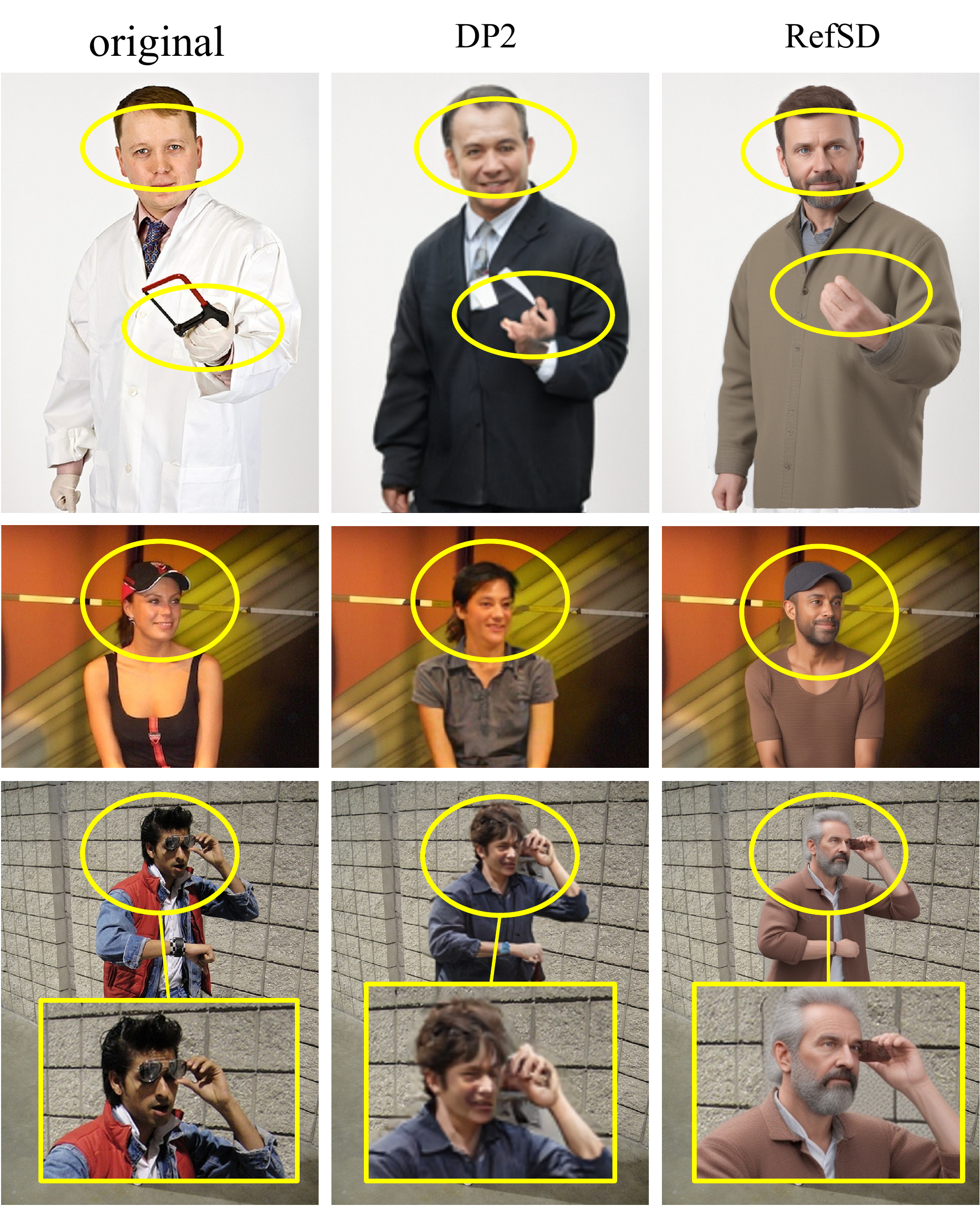}
    \caption{Comparison of DeepPrivacy2 (DP2) and RefSD (ours). RefSD produces more accurate and realistic humans, especially in fine-grain details like face and gesture generation.}
    \label{fig:appendix_refsd_dp2_comparison}
\end{figure}

Further comparisons between RefSD, regular Stable Diffusion (SD), and DeepPrivacy2 (DP2) are presented in Fig.~\ref{fig:appendix_refsd_comparison}, complementing the comparisons in Fig.~\ref{fig:appendix_refsd_dp2_comparison}, which focus specifically on DP2 vs. RefSD.

Stable Diffusion (SD), lacking posture preservation constraints, generates individuals without accounting for the original pose when provided with a mask. This results in a significant degradation of the utility and context of the image, as the generated content no longer aligns with the scene's original semantics. In contrast, DP2 employs dense pose estimation to preserve the pose, which partially retains the original context. However, its reliance on GANs limits its capability to produce realistic images, as GANs lack the generation control offered by newer diffusion-based methods. Furthermore, the realism and fine-grained detail in DP2 outputs are compromised due to the outdated nature of GAN-based architectures.

RefSD addresses these challenges by leveraging 3D rendered avatars to preserve posture while maintaining control over the generation process. This approach ensures superior preservation of fine-grained features, resulting in more realistic and contextually aligned outputs. As shown in Fig.~\ref{fig:appendix_refsd_comparison} and Fig.~\ref{fig:appendix_refsd_dp2_comparison}, RefSD excels in rendering intricate details, such as facial features and hand gestures, further enhancing the quality and utility of the generated images.


\section{Human Perception Evaluations ($\phi$)}
\label{appendix:phi}

This section presents the results of the human perception evaluations conducted on HumanGenAI, covering four key evaluations: Prompt Complexity ($\phi_{\text{A}}$), Individual Attribute — Face ($\phi_{\text{B}}$), Fine-Grain Attribute Translation ($\phi_{\text{C}}$), and Individual Attribute — Full Body ($\phi_{\text{D}}$). For each evaluation, we provide a detailed breakdown of the categories and attributes, present all results for each category, and include sample RefSD-generated images.

\subsection{Prompt Complexity ($\phi_{\text{A}}$)}
\label{appendix:phi_A}

\begin{figure}
    \centering
    \includegraphics[width=1\linewidth]{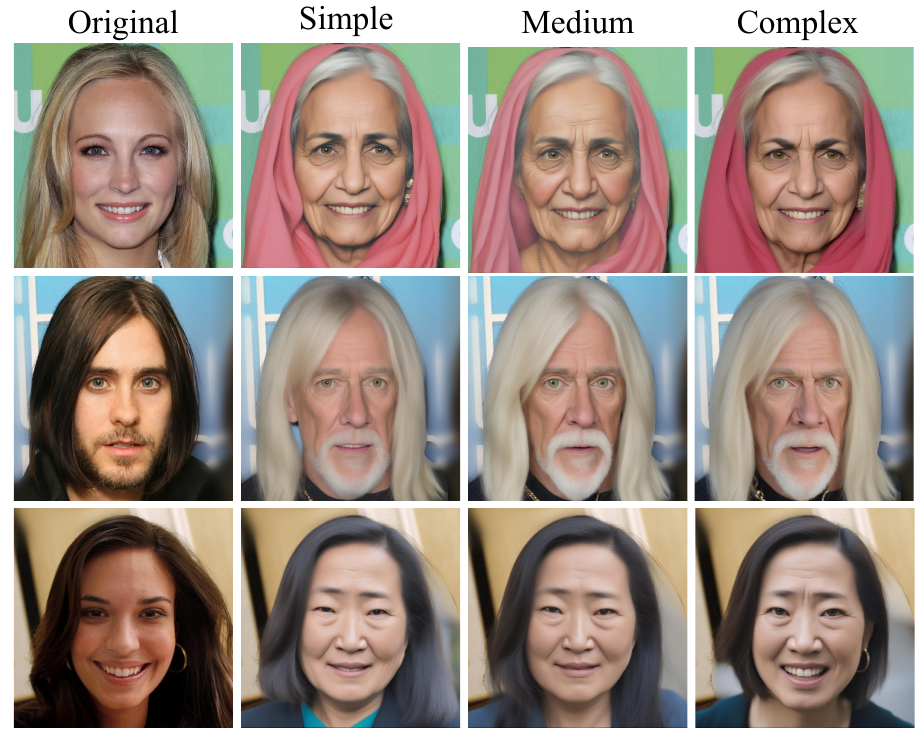}
    \caption{Example generated images showing the impact of prompt complexity ($\phi_{\text{A}}$) using simple, medium, and complex prompt types. The top row was generated with prompt attributes: {\color{AgeColor}64-year-old}, {\color{EthnicityColor}Middle Eastern}, {\color{GenderColor}Female}, {\color{FaceAttrColor}Blonde Hair}, {\color{EmotionColor}Angry}. The second row was generated with prompt attributes: {\color{AgeColor}54-year-old}, {\color{EthnicityColor}White}, {\color{GenderColor}Male}, {\color{FaceAttrColor}Necklace}, {\color{EmotionColor}Surprised}.
    The last row was generated with prompt attributes: 
    {\color{AgeColor}44-year-old}, {\color{EthnicityColor}East Asian}, {\color{GenderColor}Female}, {\color{FaceAttrColor}Straight Hair}, {\color{EmotionColor}Disgusted}.
    }
    \label{fig:a2_examples_appendix}
\end{figure}

\begin{figure*}[htp]
    \centering
    \includegraphics[width=0.8\linewidth]{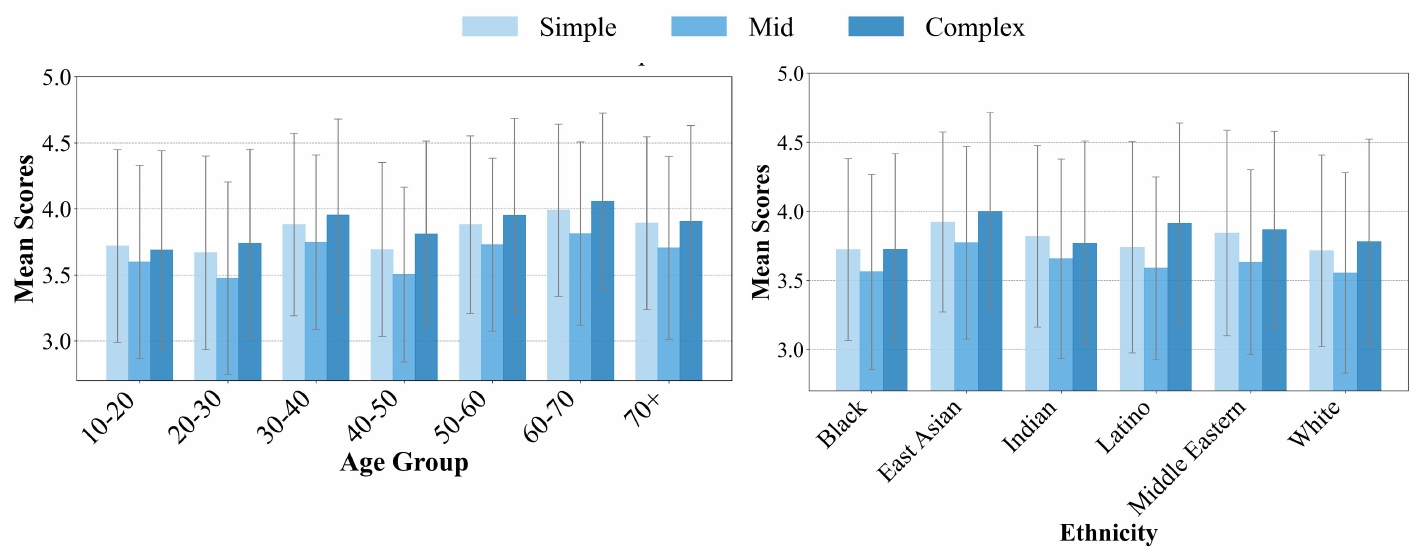}
    \caption{Mean annotator scores for Prompt Complexity ($\phi_{\text{A}}$) for \textbf{Age} (left) and \textbf{Ethnicity} (right).}
    \label{fig:a2_age_eth_appendix}
\end{figure*}

\begin{figure*}[htp]
    \centering
    \includegraphics[width=0.8\linewidth]{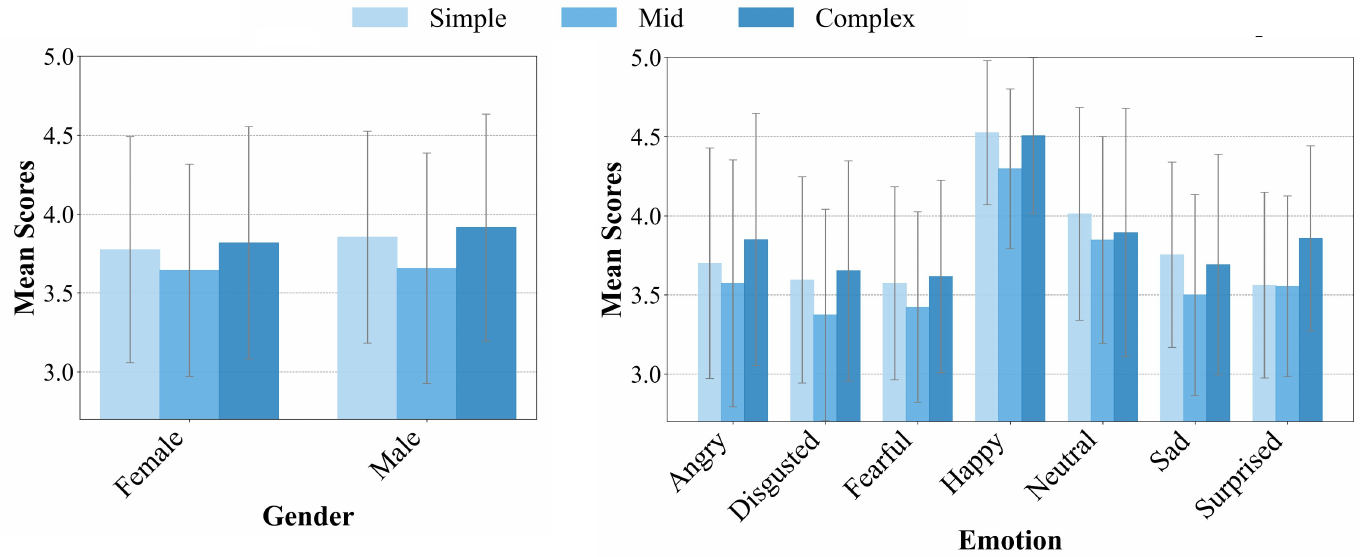}
    \caption{Mean annotator scores for Prompt Complexity ($\phi_{\text{A}}$) for \textbf{Gender} (left) and \textbf{Emotion} (right).}
    \label{fig:a2_gend_emt_appendix}
\end{figure*}

\begin{figure*}[htp]
    \centering
    \includegraphics[width=1.0\linewidth]{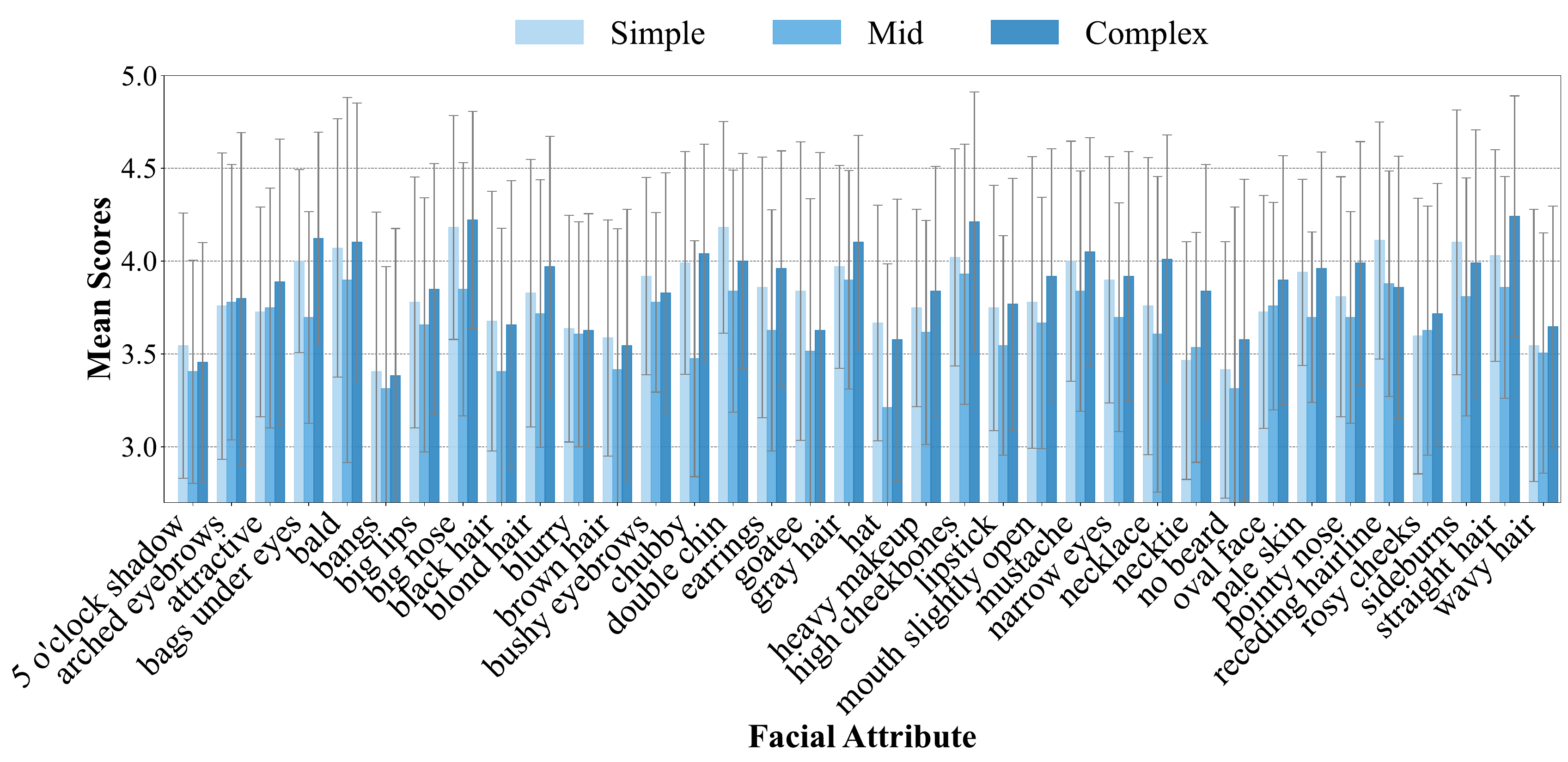}
    \caption{Mean annotator scores for Prompt Complexity ($\phi_{\text{A}}$) for \textbf{Facial Attributes}.}
    \label{fig:a2_face_attr_appendix}
\end{figure*}

This evaluation assesses the effect of prompt complexity on face generation by testing simple, medium, and complex prompts with the same attributes but varying additional details. Figs.~\ref{fig:a2_age_eth_appendix} to~\ref{fig:a2_face_attr_appendix} show the mean human evaluation scores per attribute across all categories. The trend observed in the main paper continues, where complex prompts receive the highest scores. However, these scores are not significantly higher than those for simple prompts. In many  Medium prompts consistently receive the lowest scores among all three levels of complexity. The overall scores across attributes in ages, ethnicities, and genders are mostly constant. Notably, for emotion, `happy' has noticeably higher mean scores, indicating it is the most well and consistently generated or visually satisfying emotion. Face attributes also show more fluctuation. Considering these prompts combine five attributes, we find that ethnicities and gender are best and consistently represented, followed by age, then emotion, and finally face attributes. Fig.~\ref{fig:a2_examples_appendix} shows select examples from all three prompt types. The visual differences are subtle, emphasizing the proximity in scores for all three prompt types. However, both visually and according to mean scores, complex prompt images show a slight human preference.

\textbf{Categories/Attributes for $\phi_{\text{A}}$.} This evaluation considered only face images and 5 attribute categories: Age (7 groups), ethnicity (7), gender (2), facial attributes (36), and emotions (7). The Face attributes are taken from CelebA, Ethnicity from FairFace, and Emotions from RAF-DB. These are detailed below. 

\medskip

\begin{itemize}
    \item \textbf{Gender:} \\
    \texttt{`Male', `Female'}
    \item \textbf{Age:} \\
    \texttt{`10-20', `20-30', `30-40', `40-50', `50-60', `60-70', `70+'}
    \item \textbf{Ethnicity:} \\
    \texttt{`White', `South East Asian', `Indian', `East Asian', `Middle Eastern', `Black', `Latino'}
    \item \textbf{Emotion:} \\
    \texttt{`Angry', `Neutral', `Sad', `Happy', `Fearful', `Surprised', `Disgusted'}
    \item \textbf{Face Attributes:} \\
    \texttt{`no beard', `goatee', `bald', `necklace', `5 o'clock shadow', `attractive', `chubby', `lipstick', `pale skin', `earrings', `bags under eyes', `arched eyebrows', `sideburns', `necktie', `narrow eyes', `mustache', `black hair', `wavy hair', `bangs', `big nose', `gray hair', `blurry', `big lips', `pointy nose', `high cheekbones', `receding hairline', `bushy eyebrows', `straight hair', `mouth slightly open', `oval face', `rosy cheeks', `heavy makeup', `brown hair', `double chin', `blond hair', `hat'}
    
\end{itemize}

\subsection{Individual Attribute -- Face ($\phi_{\text{B}}$)}
\label{appendix:phi_B}

\begin{figure*}[htp]
    \centering
    \includegraphics[width=1\linewidth]{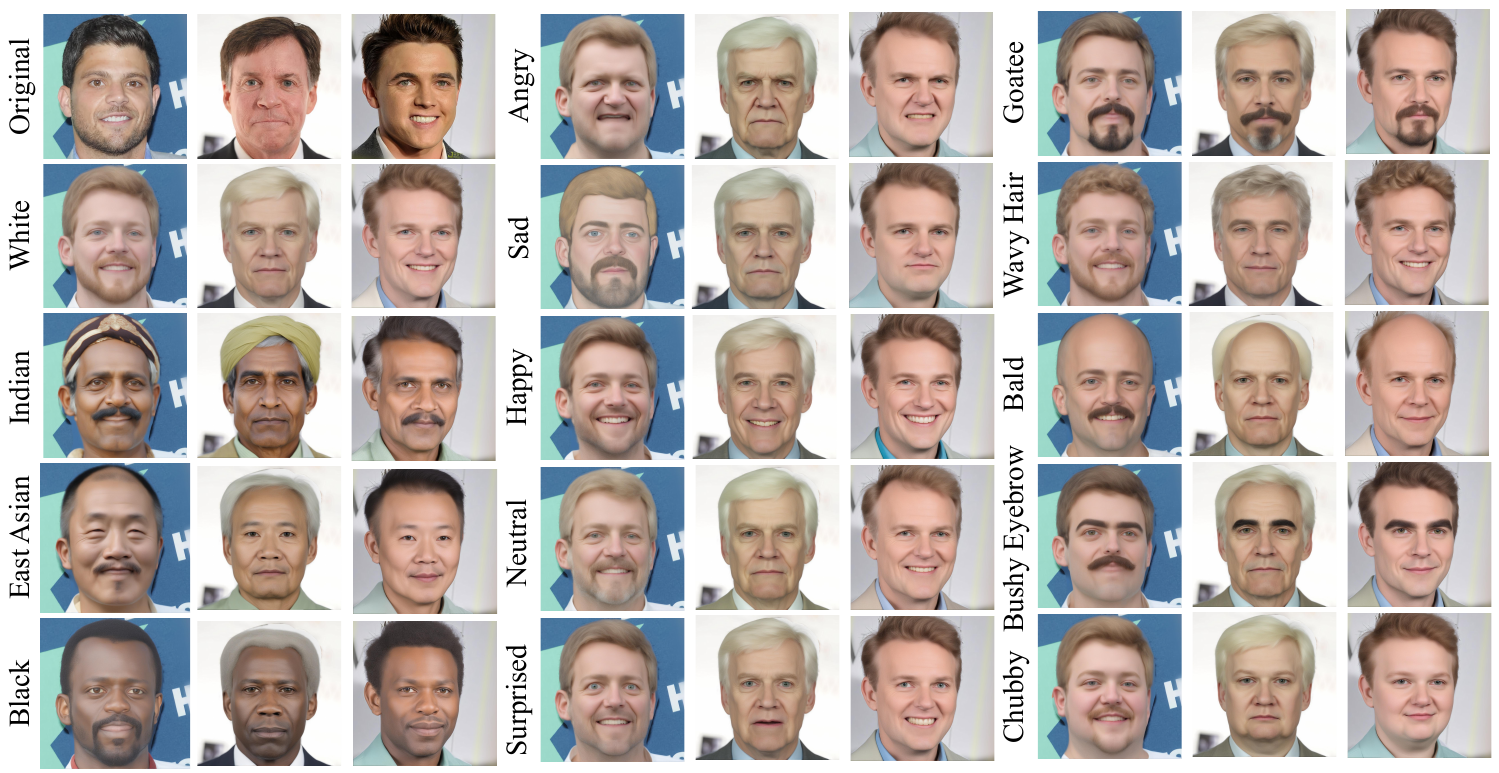}
    \caption{Example RefSD pseudonymized images for $\phi_{\text{B}}$ showcasing select ethnicities, emotions, and facial attributes. Original/source images are shown in top left.}
    \label{fig:phi_2_images}
\end{figure*}

This evaluation uses a basic prompt type and evaluates the presence and generation of a single attribute at a time. We detail the attribute categories below. Fig.~\ref{fig:phi_2} in main paper shows the human mean scores and standard deviations across all 50 attributes, separated into three categories: ethnicity, emotion, and facial attributes. Ethnicity has the highest mean and very high scores. For emotion, only `happy' and `angry' have high mean scores, with the rest being lower. The facial attributes have more diverse scores, with some attributes like `lipstick' and `goatee' having high scores, while others like `5 o'clock shadow' and `no beard' are lower. Fig.~\ref{fig:phi_2_images} show some visual examples of generated images for select ethnicities, emotions, and facial attributes.

\textbf{Categories/Attributes for $\phi_{\text{B}}$.} This evaluation considered Face images only, and 3 attribute categories: Ethnicity (7), Emotion (7), and Face attributes (36). These are the same as $\phi_{\text{A}}$ described above Sec.~\ref{appendix:phi_A}. Hence, a total of 50 unique attributes were considered.




\subsection{Fine-Grain Attribute Translation ($\phi_{\text{C}}$)}
\label{appendix:phi_C}

\begin{figure*}[h]
    \centering
    \includegraphics[width=0.8\linewidth]{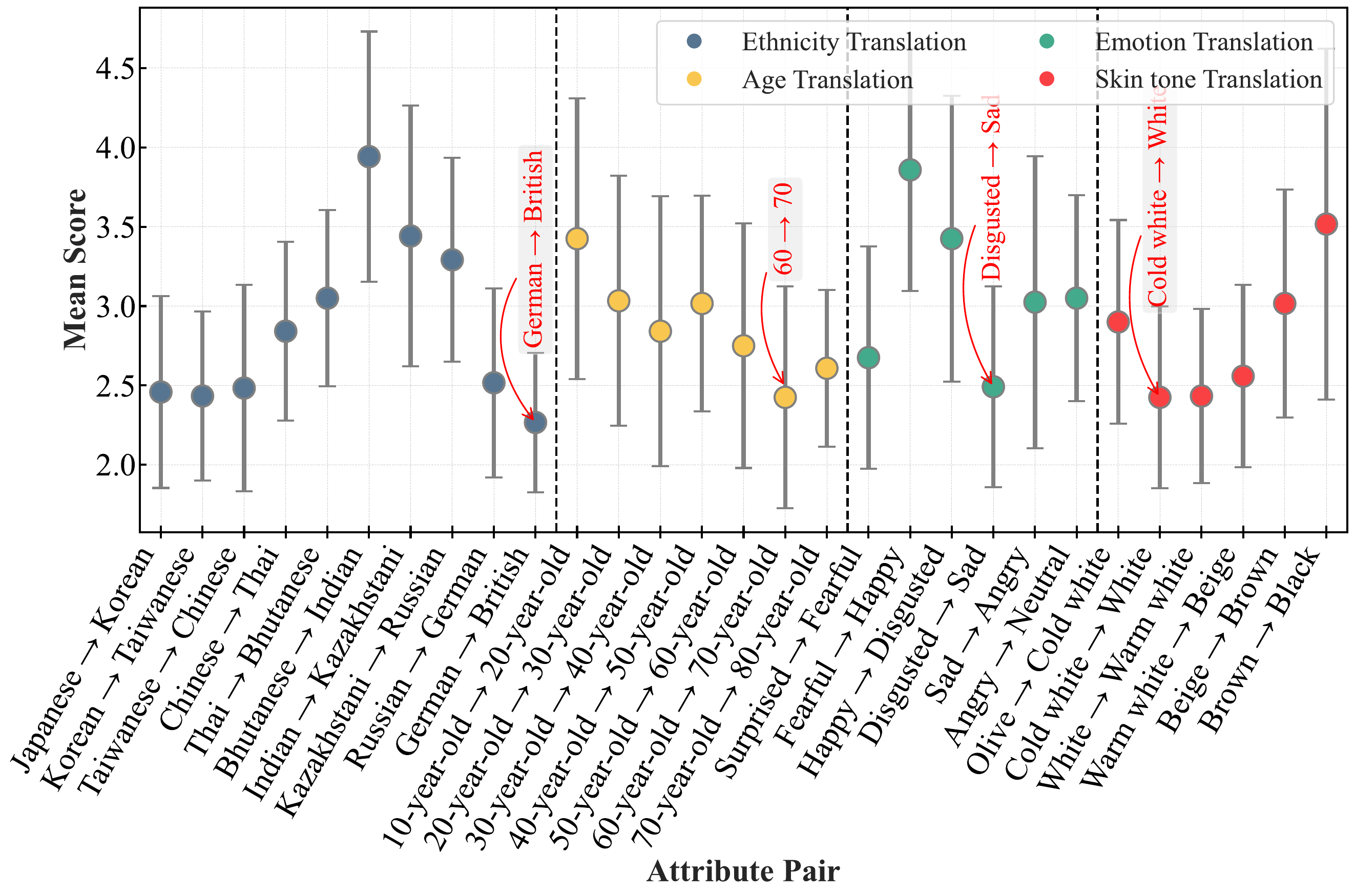}
    \caption{Mean annotation scores for \textbf{Fine-Grained Attribute Translation ($\phi_{\text{C}}$)} across Ethnicity, Emotion, Age, and Skin tone groups. Highlighted are the pairs with the lowest mean scores.} \label{fig:phi_3_appendix}
\end{figure*}

\begin{figure*}[h]
    \centering
    \includegraphics[width=0.95\linewidth]{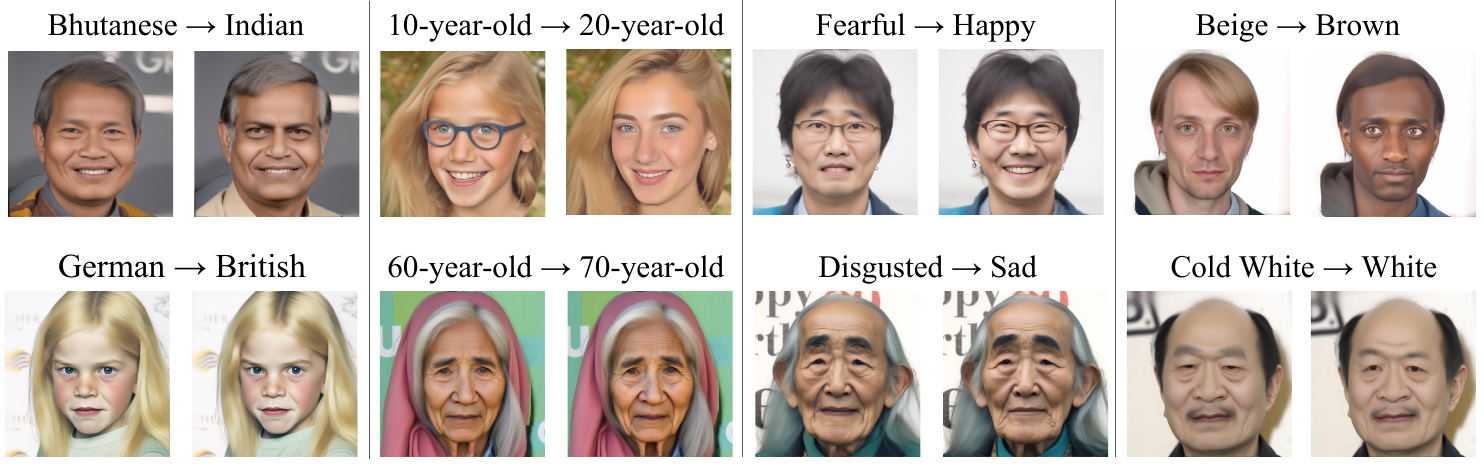}
    \caption{Example generated images for $\phi_{\text{C}}$. The top row shows the transition pair with the highest mean score, while the bottom row shows the pair with the lowest mean score (as indicated in Figure~\ref{fig:phi_3_appendix}). The image pairs, from left to right, represent transitions in ethnicity, age, emotion, and skin tone, respectively.}
    \label{fig:phi_3_examples}
\end{figure*}

Fine-grain attribute translation explores attributes that are similar in nature to determine if RefSD can generate visually distinct or separable images. We consider our proposed attributes, which are described below. Fig.~\ref{fig:phi_3_appendix} (Fig.~\ref{fig:phi_3} from main paper, shown in higher resolution) shows the human mean scores across all four categories (ethnicity, age, emotion, and skin tones) for each translation pair. The top mean scores for each category are Bhutanese $\rightarrow$ Indian for Ethnicity, 10-year-old $\rightarrow$ 20-year-old for Age, Fearful $\rightarrow$ Happy for Emotion, and Beige $\rightarrow$ Brown for Skin Tones. These pairs exhibit visually distinct components; for instance, Indians are stereotypically represented, the transition from 10 to 20 years shows a larger visual difference, happy is the best-generated and pronounced emotion, and the translation from beige to brown is more noticeable compared to others. We also highlight the lowest scored pairs, where there was very minimal difference in the images. We showcase generated examples for these trends in Fig.~\ref{fig:phi_3_examples}.

\textbf{Categories/Attributes for $\phi_{\text{C}}$.} This evaluation considers 4 attribute categories; Ethnicity (10 pairs), Age (7 pairs), Emotion (6 pairs), and Skin tones (6 pairs). These are detailed below.

\begin{itemize}
    \item \textbf{Ethnicity:} \\
    \texttt{Japanese $\rightarrow$ Korean, Korean $\rightarrow$ Taiwanese, Taiwanese $\rightarrow$ Chinese, Chinese $\rightarrow$ Thai, Thai $\rightarrow$ Bhutanese, Bhutanese $\rightarrow$ Indian, Indian $\rightarrow$ Kazakhstani, Kazakhstani $\rightarrow$ Russian, Russian $\rightarrow$ German, German $\rightarrow$ British}  
    \item \textbf{Age:} \\
    \texttt{10-year-old $\rightarrow$ 20-year-old, 20-year-old $\rightarrow$ 30-year-old, 30-year-old $\rightarrow$ 40-year-old, 40-year-old $\rightarrow$ 50-year-old, 50-year-old $\rightarrow$ 60-year-old, 60-year-old $\rightarrow$ 70-year-old, 70-year-old $\rightarrow$ 80-year-old}
    \item \textbf{Emotion:} \\
    \texttt{Surprised $\rightarrow$ Fearful, Fearful $\rightarrow$ Happy, Happy $\rightarrow$ Disgusted, Disgusted $\rightarrow$ Sad, Sad $\rightarrow$ Angry, Angry $\rightarrow$ Neutral}
    \item \textbf{Skin Tones:} \texttt{Olive $\rightarrow$ Cold white, Cold white $\rightarrow$ White, White $\rightarrow$ Warm white, Warm white $\rightarrow$ Beige, Beige $\rightarrow$ Brown, Brown $\rightarrow$ Black}
\end{itemize} 

\subsection{Individual Attribute -- Full Body ($\phi_{\text{D}}$)}
\label{appendix:phi_D}

\begin{figure*}[h]
    \centering
    \includegraphics[width=0.9\linewidth]{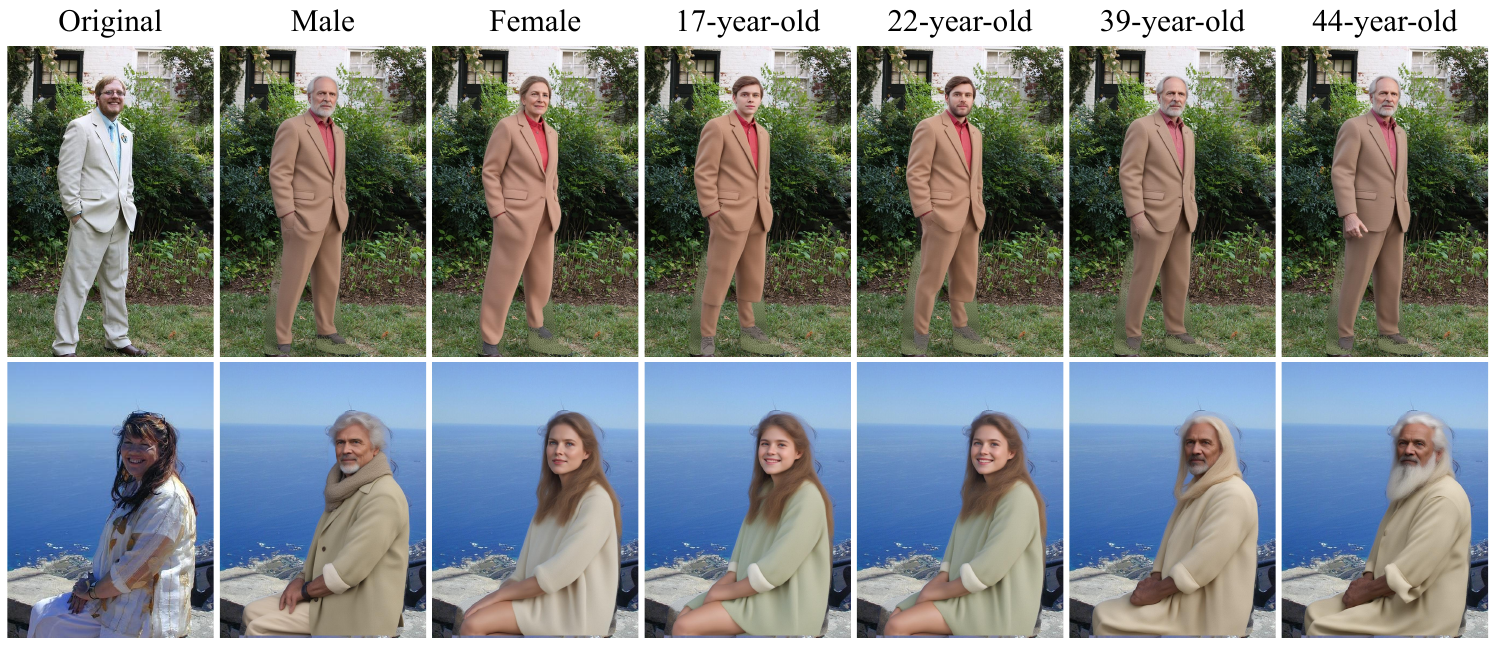}
    \caption{Example synthesized images for \textbf{$\phi_{\text{D}}$}, illustrating select \textbf{Gender} and \textbf{Age} using basic prompts.}
    \label{fig:phi5_gender_age_imgs}
\end{figure*}

\begin{figure*}[h]
    \centering
    \includegraphics[width=0.6\linewidth]{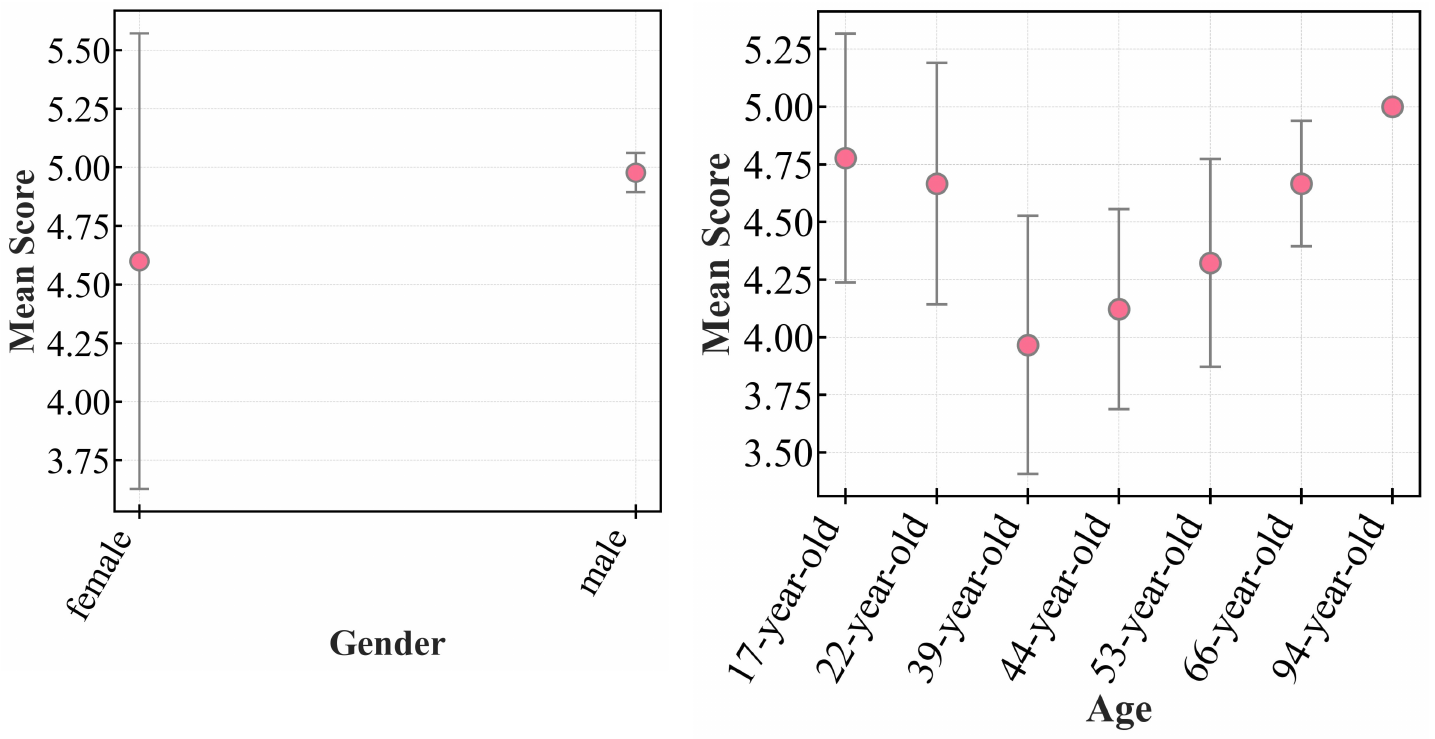}
    \caption{The mean scores given by annotators for \textbf{$\phi_{\text{D}}$} for \textbf{Gender} (left) and \textbf{Age} (right).}
    \label{fig:phi5_gender_age}
\end{figure*}

\begin{figure*}[h]
    \centering
    \includegraphics[width=0.9\linewidth]{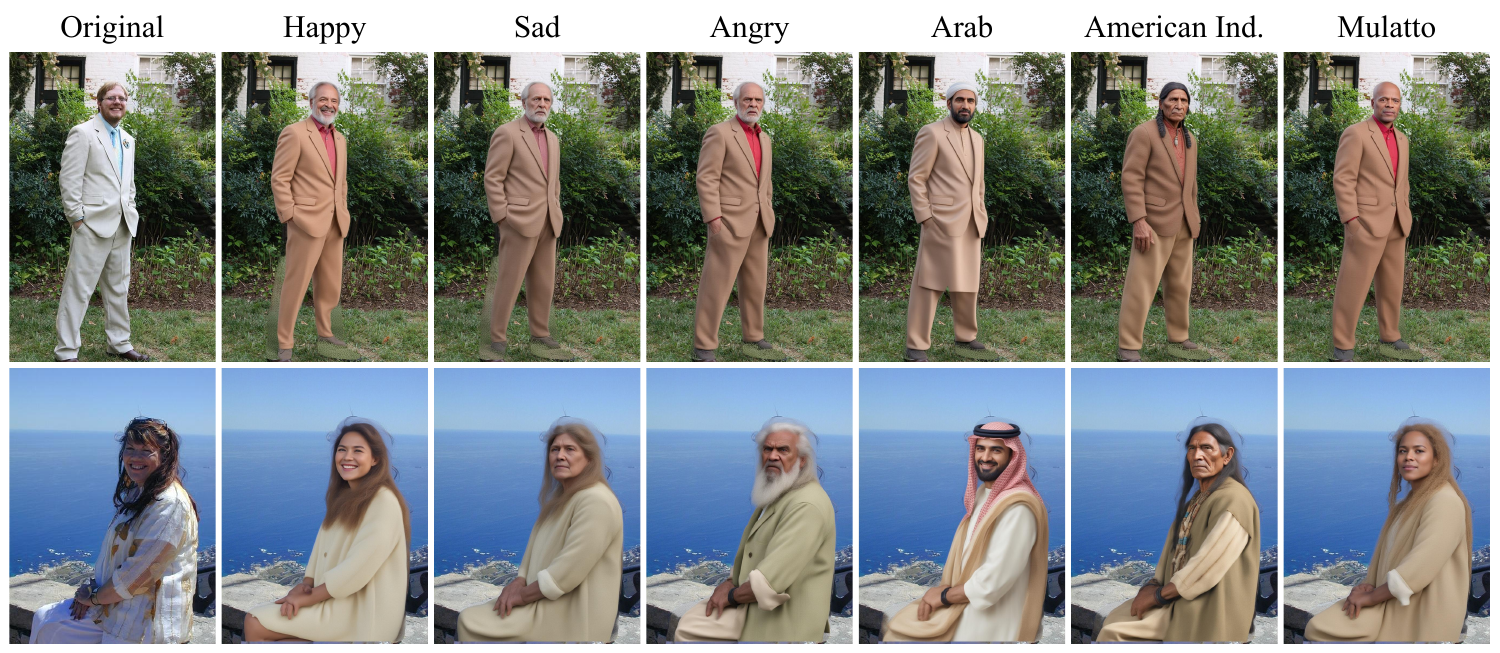}
    \caption{Example synthesized images for \textbf{$\phi_{\text{D}}$}, illustrating select \textbf{Emotions} and \textbf{Ethnicities} using basic prompts. We show additional ethnicities compared to Fig.~\ref{fig:phi_2_images}, including Arab, American Indian, and Mulatto.}
    \label{fig:phi5_emotion_ethn_imgs}
\end{figure*}

\begin{figure*}[h]
    \centering
    \includegraphics[width=0.7\linewidth]{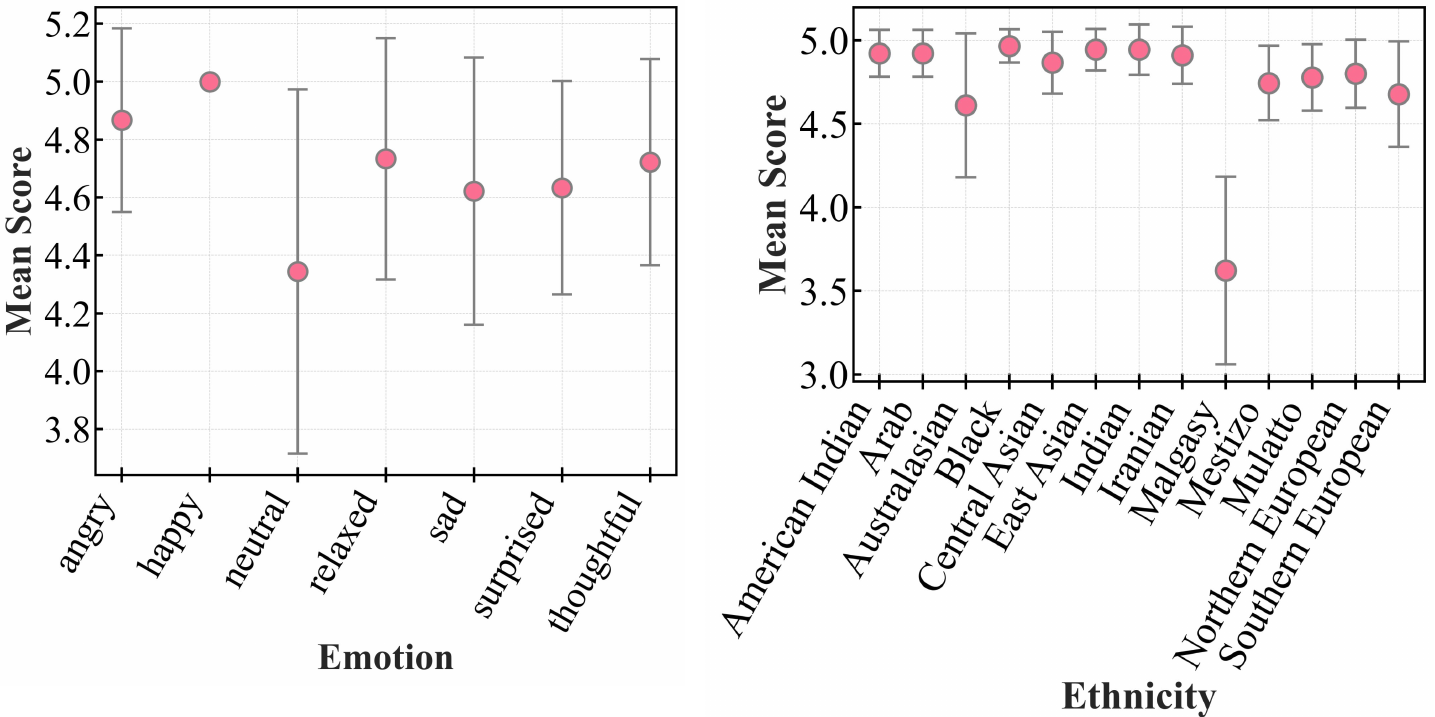}
    \caption{The mean scores given by annotators for \textbf{$\phi_{\text{D}}$} for \textbf{Emotion} (left) and \textbf{Ethnicity} (right).}
    \label{fig:phi5_emotion_ethn}
\end{figure*}

\begin{figure*}[h]
    \centering
    \includegraphics[width=0.9\linewidth]{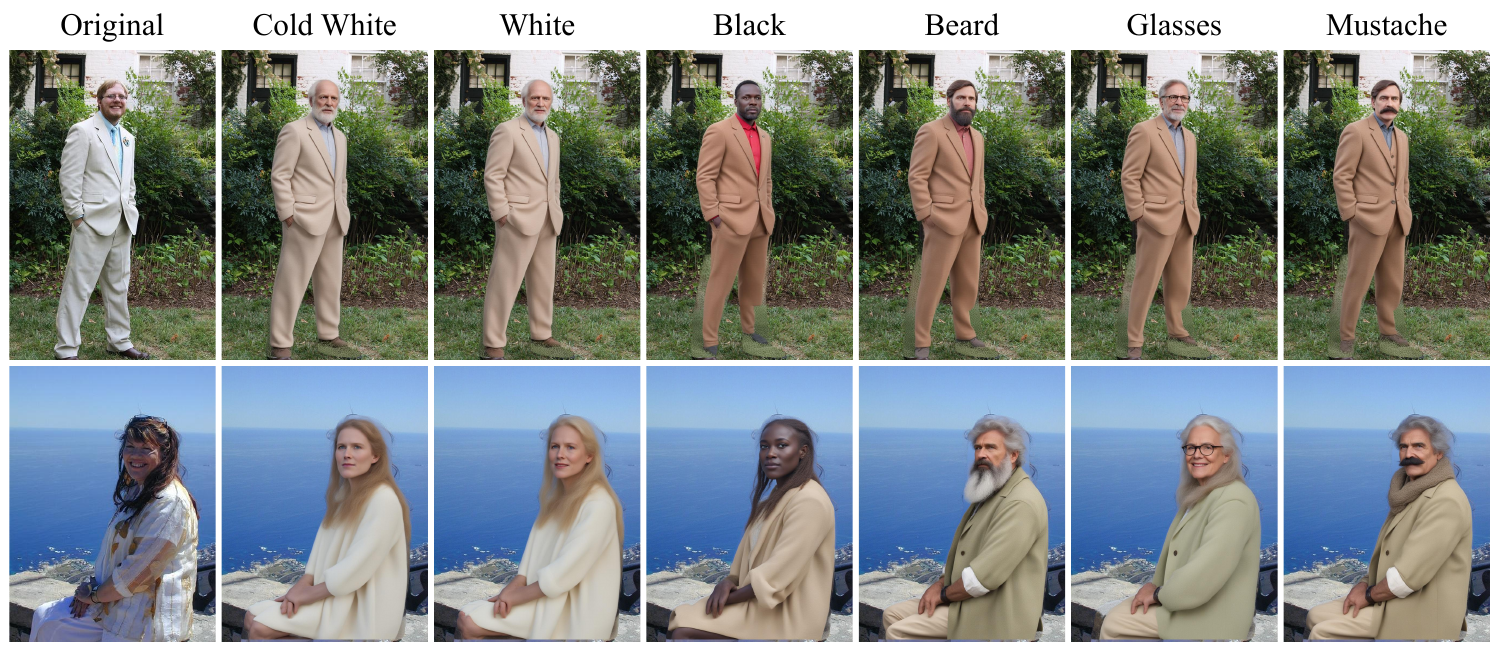}
    \caption{Example synthesized images for \textbf{$\phi_{\text{D}}$}, illustrating select \textbf{Skin tone} and \textbf{Facial features} using basic prompts. }
    \label{fig:phi5_skin_face_imgs}
\end{figure*}

\begin{figure*}[h]
    \centering
    \includegraphics[width=0.7\linewidth]{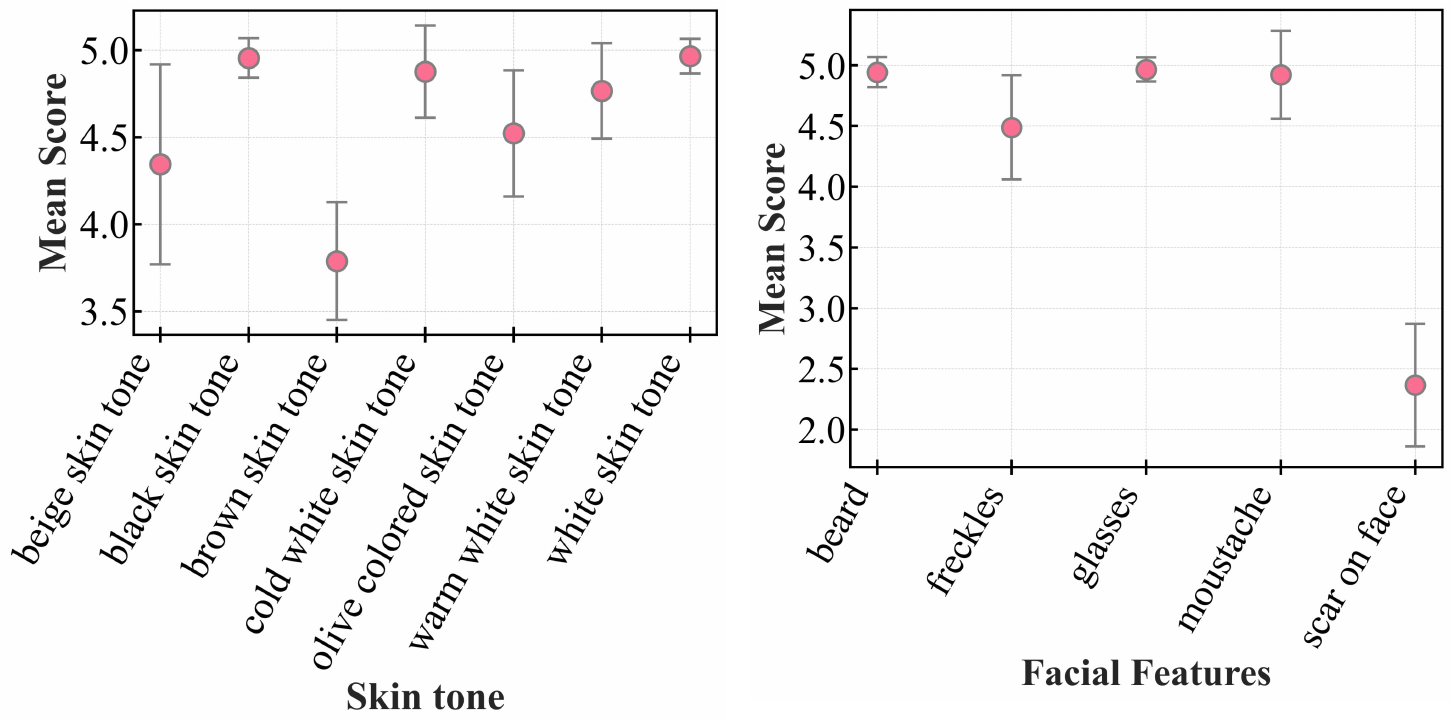}
    \caption{The mean scores given by annotators for \textbf{$\phi_{\text{D}}$} for \textbf{Skin tone} (left) and \textbf{Face features} (right).}
    \label{fig:phi5_skin_face}
\end{figure*}

\begin{figure*}[h]
    \centering
    \includegraphics[width=0.85\linewidth]{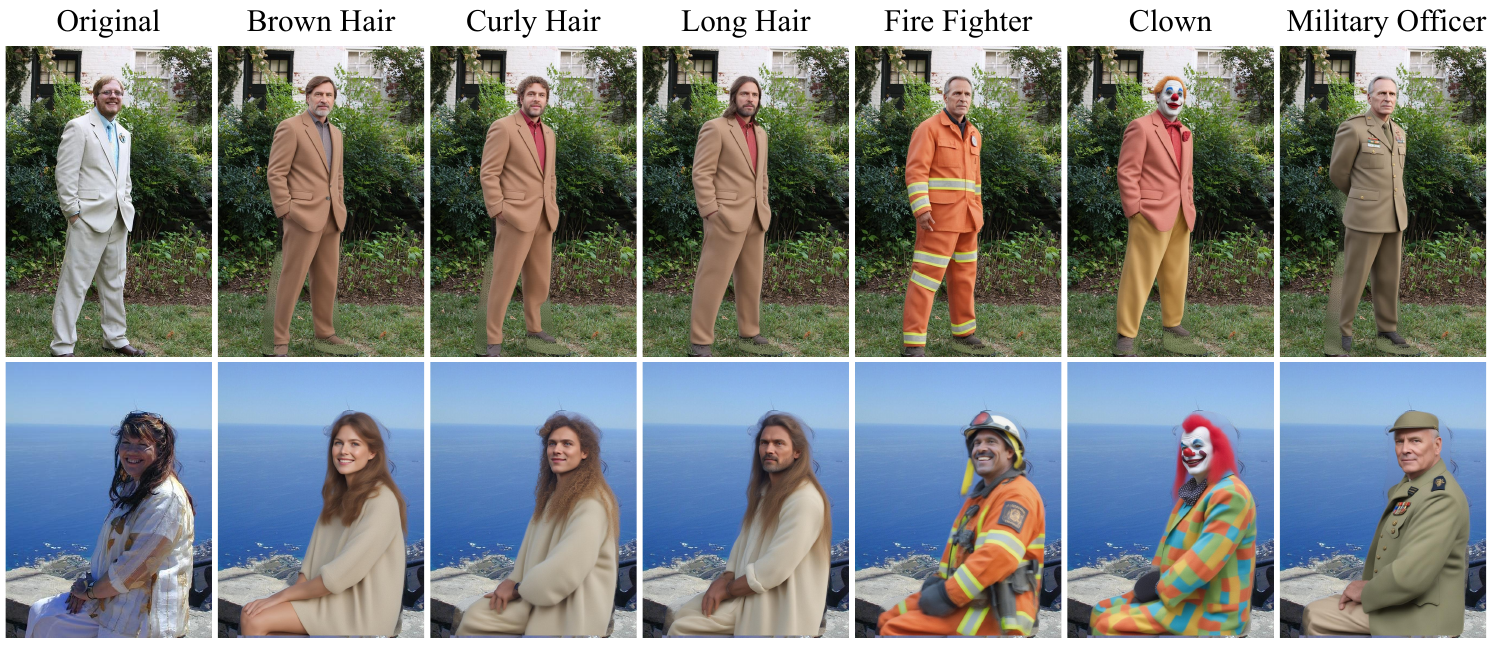}
    \caption{Example synthesized images for \textbf{$\phi_{\text{D}}$}, illustrating select \textbf{Hair Color \& Style} and \textbf{Occupation} using basic prompts. }
    \label{fig:phi5_hair_occupation_imgs}
\end{figure*}

\begin{figure*}[h]
    \centering
    \includegraphics[width=0.65\linewidth]{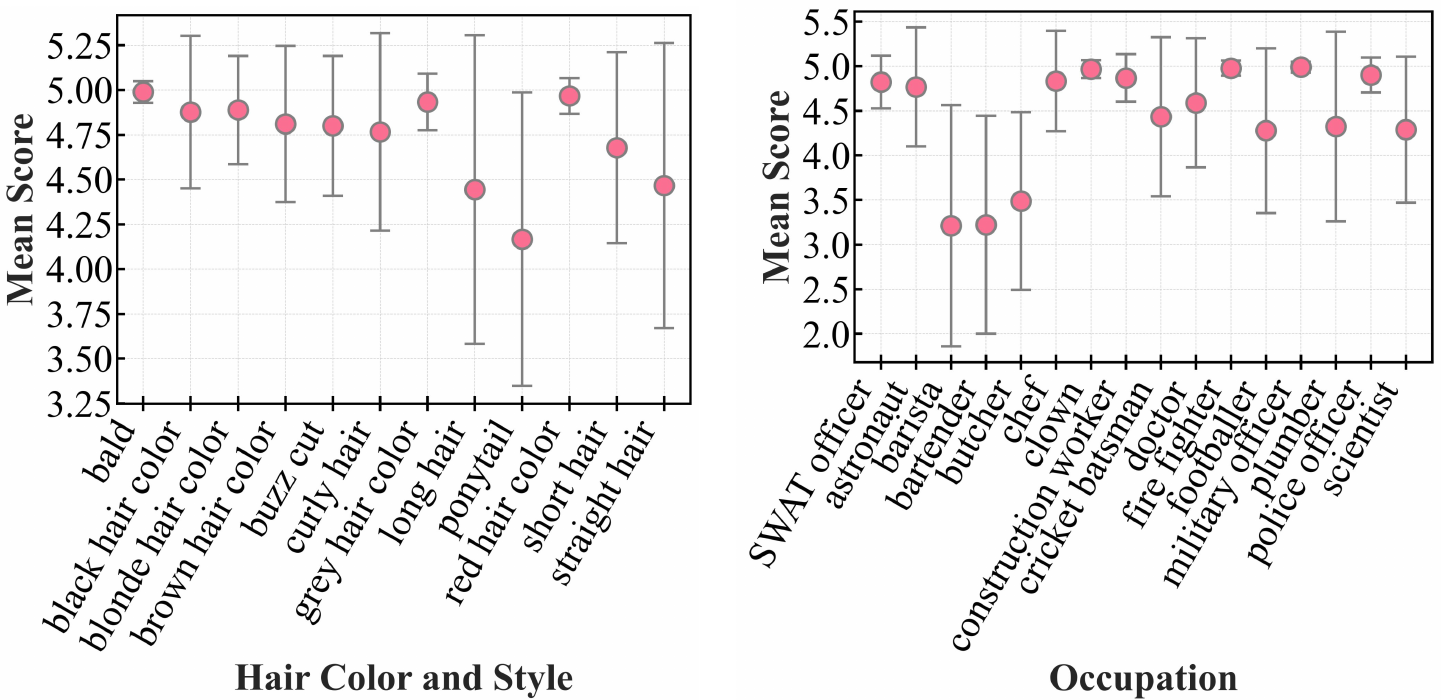}
    \caption{The mean scores given by annotators for \textbf{$\phi_{\text{D}}$} for \textbf{Hair color \& style} (left) and \textbf{Occupation} (right).}
    \label{fig:phi5_hair_occupation}
\end{figure*}

\begin{figure*}[h]
    \centering
    \includegraphics[width=0.85\linewidth]{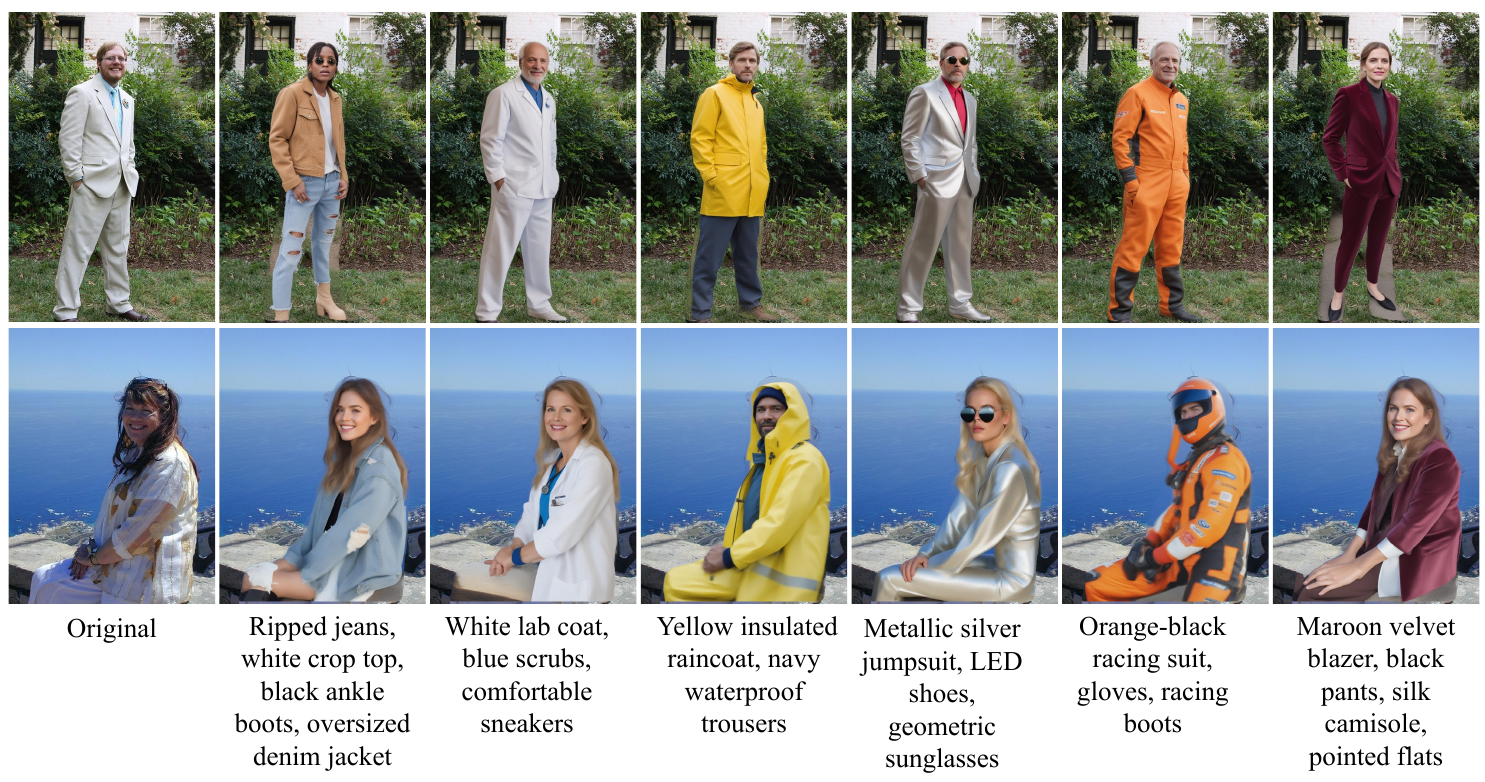}
    \caption{Example synthesized images for \textbf{$\phi_{\text{D}}$}, illustrating select \textbf{Clothing} using basic prompts. }
    \label{fig:phi5_clothing_imgs}
\end{figure*}

\begin{figure*}[h]
    \centering
    \includegraphics[width=1\linewidth]{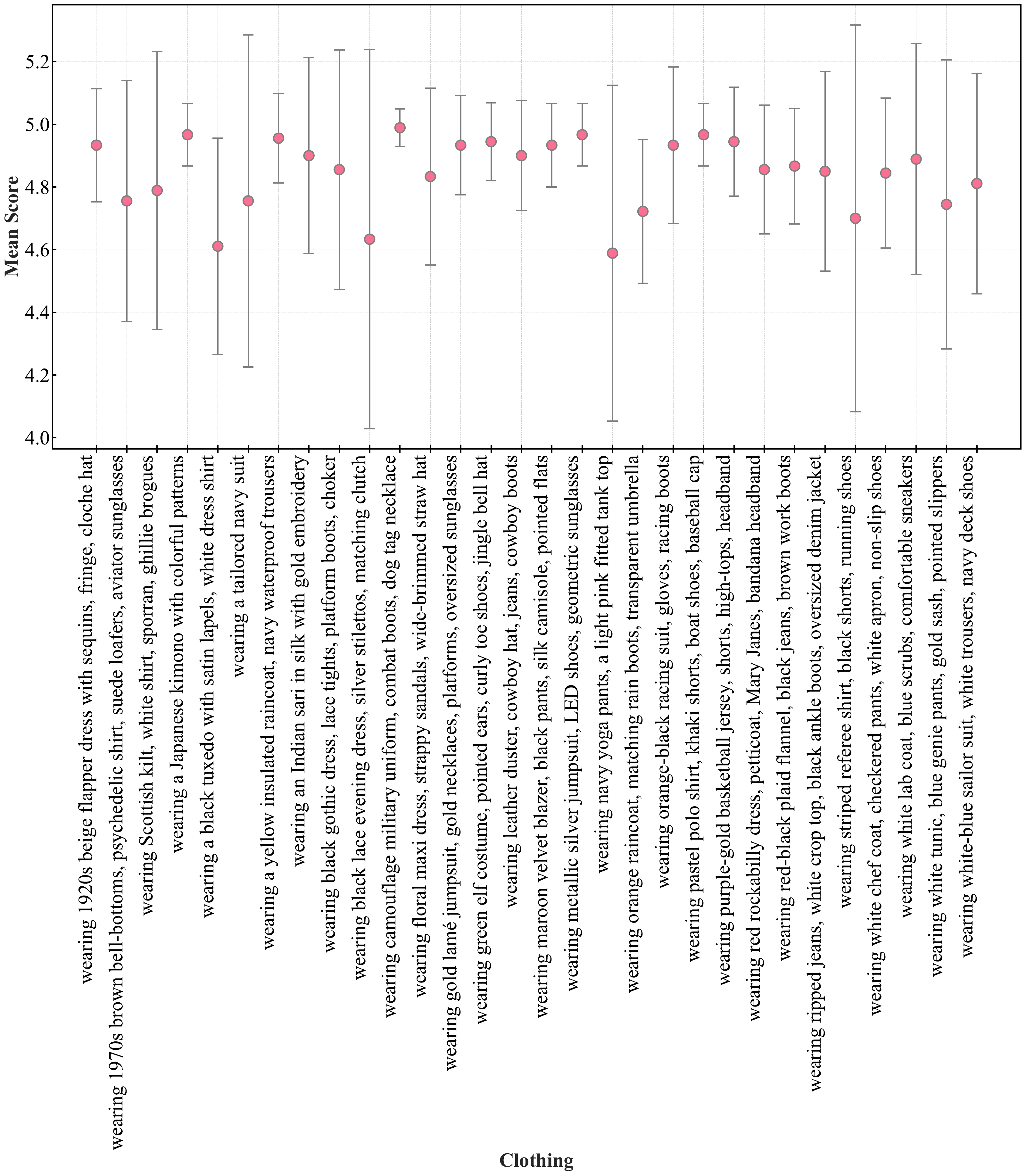}
    \caption{The mean scores given by annotators for \textbf{$\phi_{\text{D}}$} for \textbf{Clothing}.}
    \label{fig:phi5_clothing}
\end{figure*}

We follow $\phi_{\text{B}}$ but generate images for full-body humans rather than faces. To study fine-grain attributes and features, and distinguish this evaluation from $\phi_{\text{B}}$, we propose a new set of attributes, some of which are specifically categorized for full-body humans. We describe all the attributes and categories used below. Fig.~\ref{fig:phi_4} (main paper) shows the human mean scores for all attribute categories, with Figs.~\ref{fig:phi5_gender_age_imgs} to~\ref{fig:phi5_clothing} providing sample example images and individual breakdowns (mean annotation scores) for each category. These figures offer a visual indication of which individual attributes within each category are well-generated and which are not, according to human evaluators.

\textbf{Categories/Attributes for $\phi_{\text{D}}$.} This evaluation considers full body images, and 7 attribute categories: Gender (2), Age (7), Emotion (7), Ethnicity (13), Skin Tone (7), Face attributes/features (5), Hair color \& style (7), Clothing style (31), and Occupation (16). Hence, a total of 100 unique attributes were considered. Age and Gender are same as A.2, the remaining are detailed below.

\begin{itemize}
    \item \textbf{Gender:} \\
    \texttt{`Male', `Female'}
    \item \textbf{Age:} \\
    \texttt{`17-year-old', `22-year-old', `39-year-old', `44-year-old', `53-year-old',
    `66-year-old', `94-year-old'}
    \item \textbf{Emotion:} \\
    \texttt{`Happy', `Sad', `Angry', `Surprised', `Thoughtful', `Relaxed', `Neutral'}  
    \item \textbf{Ethnicity:} \\
    \texttt{`Northern European', `Southern European', `Arab`, `Central Asian', `Indian', `Iranian', `Mulatto', `Black', `East Asian', `Malgasy', `American Indian', `Mestizo', `Australasian'}
    \item \textbf{Skin Tone:} \\
    \texttt{`white', `beige', `brown', `black', `warm white', `cold white', `olive'}
    \item \textbf{Face attr./features:} \\
    \texttt{`Freckles', `Glasses', `Beard', `Moustache', `Scar on face'}
    \item \textbf{Hair Color \& Style:} \\
    \texttt{`blonde', `brown', `black', `red', `grey', `short', `long', `curly', `straight', `ponytail', `buzz cut', `bald'}
    \item \textbf{Occupation:} \\
    \texttt{`footballer', `chef', `police officer', `fire fighter', `astronaut', `construction worker', `clown', `barista', `bartender', `butcher', `doctor', `military officer', `scientist', `cricket batsman', `SWAT officer', `plumber'}
    \item \textbf{Clothing Style:}  \\
    \texttt{`wearing a black tuxedo with satin lapels, white dress shirt',\\`wearing ripped jeans, white crop top, black ankle boots, oversized denim jacket',\\`wearing white lab coat, blue scrubs, comfortable sneakers', \\`wearing camouflage military uniform, combat boots, dog tag necklace',\\ `wearing a yellow insulated raincoat, navy waterproof trousers',\\ `wearing 1970s brown bell-bottoms, psychedelic shirt, suede loafers, aviator sunglasses',\\ `wearing striped referee shirt, black shorts, running shoes',\\ `wearing metallic silver jumpsuit, LED shoes, geometric sunglasses',\\ `wearing navy yoga pants, a light pink fitted tank top',\\ `wearing white chef coat, checkered pants, white apron, non-slip shoes',\\ `wearing floral maxi dress, strappy sandals, wide-brimmed straw hat',\\ `wearing red-black plaid flannel, black jeans, brown work boots',\\ `wearing an Indian sari in silk with gold embroidery',\\ `wearing orange-black racing suit, gloves, racing boots', `wearing purple-gold basketball jersey, shorts, high-tops, headband',\\ `wearing green elf costume, pointed ears, curly toe shoes, jingle bell hat',\\ `wearing Scottish kilt, white shirt, sporran, ghillie brogues',\\ `wearing a tailored navy suit',\\ `wearing white-blue sailor suit, white trousers, navy deck shoes',\\ `wearing ripped jeans, white crop top, black ankle boots, oversized denim jacket',\\ `wearing maroon velvet blazer, black pants, silk camisole, pointed flats',\\ `wearing black gothic dress, lace tights, platform boots, choker',\\ `wearing leather duster, cowboy hat, jeans, cowboy boots',\\ `wearing gold lamé jumpsuit, gold necklaces, platforms, oversized sunglasses',\\ `wearing 1920s beige flapper dress with sequins, fringe, cloche hat',\\ `wearing orange raincoat, matching rain boots, transparent umbrella', `wearing pastel polo shirt, khaki shorts, boat shoes, baseball cap',\\ `wearing black lace evening dress, silver stilettos, matching clutch',\\ `wearing white tunic, blue genie pants, gold sash, pointed slippers',\\ `wearing red rockabilly dress, petticoat, Mary Janes, bandana headband'}
\end{itemize} 


\section{Utility Evaluations ($\psi$)}
\label{appendix:psi}

This section details the training process for utility evaluations of HumanGenAI, for both classification ($\psi_{\text{A}}$) and detection models ($\psi_{\text{B}}$). We outline the training configurations, categories, including model architectures, optimization parameters, data augmentation techniques, and evaluation metrics, providing a comprehensive overview of the methodology used to assess utility.

\subsection{Utility Training: Classification ($\psi_{\text{A}}$)}
\label{appendix:psi_A}

In the classification evaluation, we trained classifiers on pseudonymized datasets using their original labels to assess whether RefSD preserves label information, enabling the pseudonymization of labeled datasets for commercial use without compromising utility.

\textbf{Categories/Attributes for $\psi_{\text{A}}$.} This evaluation considers four attribute categories; emotion (7 classes), age (5 classes), gender (3 classes), and race (3 classes). The source images and labels are taken from RAF-DB~\ref{tab:rafdb1}. These are detailed below.

\medskip

\begin{itemize}
    \item \textbf{Race:} \\
    \texttt{`Caucasian', `African-American', `Asian'}
    \item \textbf{Emotion:} \\
    \texttt{`Surprise', `Fear', `Disgust', `Happiness', `Sadness', `Anger', `Neutral'}
    \item \textbf{Age:} \\
    \texttt{`0-3', `4-19', `20-39', `40-69', `70+'}
    \item \textbf{Gender} \\
    \texttt{`Male', `Female', `Unsure'}
\end{itemize} 

\textbf{Training Details.} We trained ViT-tiny and ViT-base models from PyTorch. Training was run for 100 epochs, with early stopping based on validation loss. We used the AdamW optimizer with a learning rate of 1e-4, a weight decay of 0.01, and a batch size of 256. Data augmentation techniques included center cropping, color jitter, random horizontal flip, random resize crop, random rotation, and random resizing. All experiments were performed on a single RTX 4090 GPU.

ViT-tiny and ViT-base models were trained on RAF-DB’s train set (12,271 images) using synthetic, real, combined, and pretrain-finetune (synthetic → real) configurations, with labels (emotion, ethnicity, gender, age) embedded in RefSD prompts. Evaluation used accuracy on the RAF-DB's test set (3,068 images).


\subsection{Utility Training: Detection ($\psi_{\text{B}}$)}
\label{appendix:psi_B}

To evaluate RefSD's ability to pseudonymize humans in in-the-wild images for object detection, we assess whether it preserves the original human pose and overall image content. Detectors are trained not only for humans but also for other objects in the images.

\textbf{Training Details.} The model incorporates the DINOv2-Adapter~\cite{oquab2023dinov2} as the encoder, paired with Faster R-CNN~\cite{ren2016faster} for object detection. Training was performed for 36 epochs using the AdamW optimizer with a learning rate of 0.0001, a weight decay of 0.5, and a linear learning rate scheduler with a 500-iteration warm-up.

We conducted object detection on a subset of the OpenImages~\cite{openimages} dataset, comprising approximately 75,000 images. The validation set includes 1,564 images, covering 227 instances of Human Faces and 722 instances of the Person object class. Performance was evaluated on the OpenImages validation set using standard mean Average Precision (mAP) at IoU thresholds 0.5:0.95 and mAP at IoU 0.5. Training the detector required 18 hours on an 8$\times$H100 GPU setup.


\end{document}